\documentclass[letterpaper, 10 pt, conference]{ieeeconf}  % Comment this line out if you need a4paper

\IEEEoverridecommandlockouts                              % This command is only needed if 
\pdfoutput=1
\usepackage{graphics} % for pdf, bitmapped graphics files
\usepackage{epsfig} % for postscript graphics files
\usepackage{mathptmx} % assumes new font selection scheme installed
\usepackage{times} % assumes new font selection scheme installed
\usepackage{amsmath} % assumes amsmath package installed
\usepackage{amssymb}  % assumes amsmath package installed
\usepackage{graphicx}
\usepackage{stfloats}
\usepackage{booktabs}
\usepackage{caption}
\usepackage{subcaption} 
\usepackage{multirow}

\usepackage[misc]{ifsym}
\usepackage[T1]{fontenc}
\usepackage[style=numeric, sorting=none]{biblatex}
\usepackage{hyperref}
\hypersetup{hypertex=true,
colorlinks=true,
linkcolor=blue,
anchorcolor=blue,
citecolor=blue,
urlcolor=cyan   
}

\title{\LARGE \bf
Genie Sim 3.0 : A High-Fidelity Comprehensive \\Simulation Platform for Humanoid Robot
}

\newcommand{\smallLetter}{\scalebox{0.6}{\Letter}}
\author{Chenghao Yin\textsuperscript{*}, Da Huang\textsuperscript{*}, Di Yang\textsuperscript{*}, Jichao Wang\textsuperscript{*}, Nanshu Zhao\textsuperscript{*}, Chen Xu\textsuperscript{*}, \\Wenjun Sun, Linjie Hou, Zhijun Li, Junhui Wu, Zhaobo Liu, Zhen Xiao, \\Sheng Zhang, Lei Bao, Rui Feng, Zhenquan Pang, Jiayu Li, Qian Wang\textsuperscript{\smallLetter}, Maoqing Yao\textsuperscript{\smallLetter}% <-this % stops a space
}
\IEEEaftertitletext{
    \centering
    * Equal contribution $ ^{\smallLetter}$ Corresponding authors\\
    \vspace{10pt}
    \includegraphics[width=0.9\linewidth]{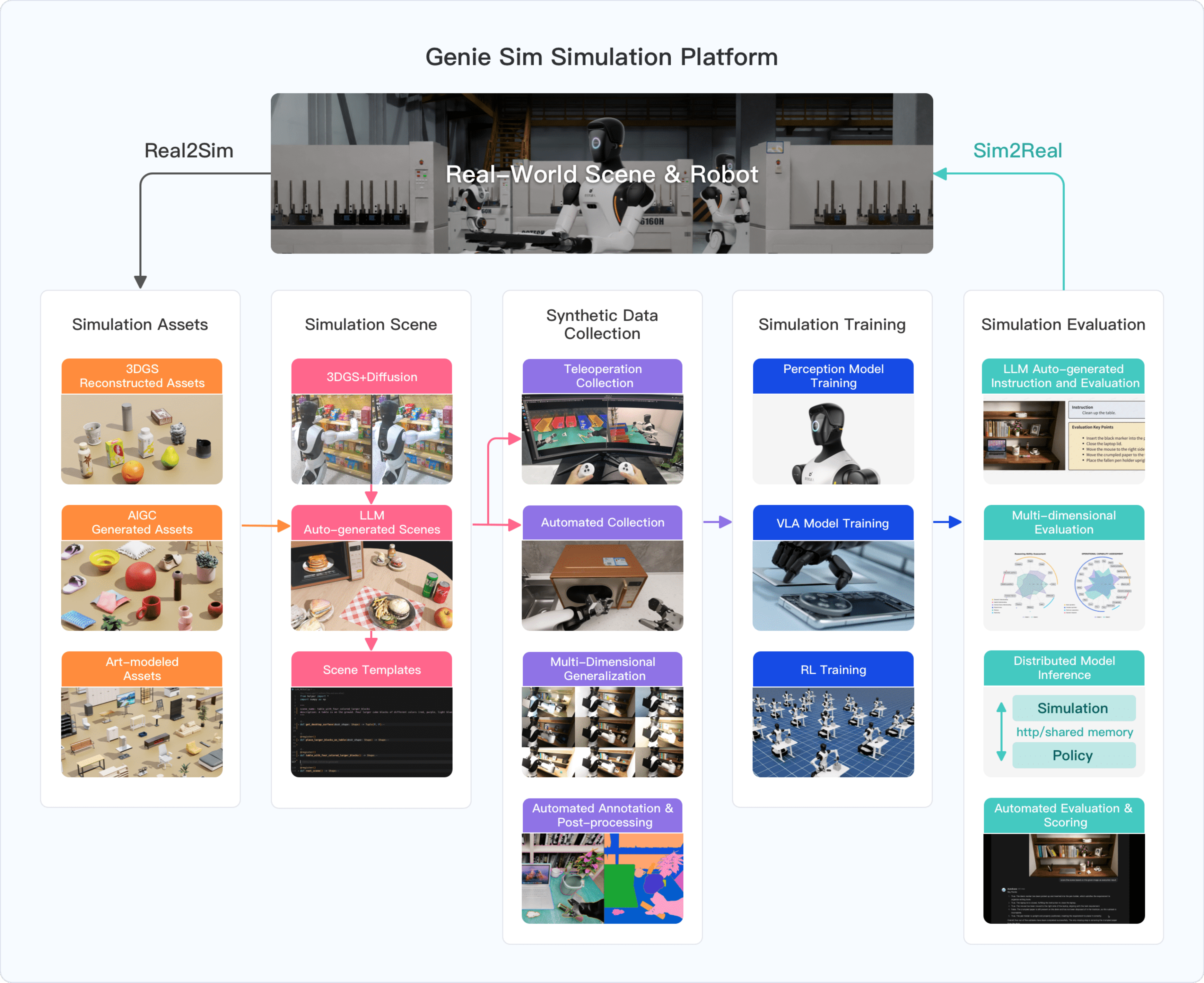}\\[6pt]
    \begin{minipage}{0.9\linewidth}
        \raggedright 
        \small 
        \textbf{Fig. 1: Overview of Genie Sim 3.0.} Genie Sim 3.0 is a full-cycle robotic simulation platform that integrates environment reconstruction, scene generalization, data collection, and automated evaluation. We plan to open-source 5,140 object assets of simulation, more than 10,000 hours of synthetic dataset and 100,000 evaluation scenarios.
    \end{minipage}
    \vspace{10pt}
}

\begin{document}
\maketitle
\thispagestyle{empty}
\pagestyle{empty}

%%%%%%%%%%%%%%%%%%%%%%%%%%%%%%%%%%%%%%%%%%%%%%%%%%%%%%%%%%%%%%%%%%%%%%%%%%%%%%%%
\setcounter{figure}{1}
\begin{abstract}

The development of robust and generalizable robot learning models is critically contingent upon the availability of large-scale, diverse training data and reliable evaluation benchmarks. Collecting data in the physical world poses prohibitive costs and scalability challenges, and prevailing simulation benchmarks frequently suffer from fragmentation, narrow scope, or insufficient fidelity to enable effective sim-to-real transfer. To address these challenges, we introduce Genie Sim 3.0, a unified simulation platform for robotic manipulation. We present Genie Sim Generator, a large language model (LLM)-powered tool that constructs high-fidelity scenes from natural language instructions. Its principal strength resides in rapid and multi-dimensional generalization, facilitating the synthesis of diverse environments to support scalable data collection and robust policy evaluation. We introduce the first benchmark that pioneers the application of LLM for automated evaluation. It leverages LLM to mass-generate evaluation scenarios and employs Vision-Language Model (VLM) to establish an automated assessment pipeline. We also release an open-source dataset comprising more than 10,000 hours of synthetic data across over 200 tasks. Through systematic experimentation, we validate the robust zero-shot sim-to-real transfer capability of our open-source dataset, demonstrating that synthetic data can server as an effective substitute for real-world data under controlled conditions for scalable policy training. For code and dataset details, please refer to: \url{https://github.com/AgibotTech/genie_sim}.

\end{abstract}

%%%%%%%%%%%%%%%%%%%%%%%%%%%%%%%%%%%%%%%%%%%%%%%%%%%%%%%%%%%%%%%%%%%%%%%%%%%%%%%%
\section{INTRODUCTION}
The advancement of robotic manipulation is increasingly supported by vision-language-action (VLA) models, which equip robots with the ability to interpret natural language instructions and execute corresponding physical actions \cite{Chi-RSS-23} \cite{kim24openvla}. This capability allows robots to perform a wider range of tasks in unstructured environments, moving closer to human-like adaptive operation  \cite{shi2025diversity}. However, the development of robust and generalizable VLA models relies heavily on access to large-scale, high-quality datasets of robot interactions, which are resource-intensive and time-consuming to collect in the real physical world \cite{jiang2024dexmimicen}\cite{chenobject}. Furthermore, systematically evaluating and iterating these models requires a consistent, reproducible and scalable testing environment, but the current evaluation approaches on real devices often fail to meet this requirement \cite{lyu2024scissorbotlearninggeneralizablescissor}.

A high-fidelity simulation platform is a practical and efficient approach to provide large-scale, cost-effective and low sim-to-real gap synthetic dataset and benchmark for VLA training and evaluation \cite{9158349}\cite{10610566}. A well-designed simulation platform can provide the necessary scale, diversity, and repeatability for both training and evaluation, significantly accelerating the development cycle of VLA-driven robotic systems \cite{jia2025discoverseefficientrobotsimulation}\cite{deng2025graspvla}. However, scaling this process encounters three fundamental bottlenecks. First, creating high-fidelity simulation environments that accurately reflect the complexity of the real world traditionally requires significant manual efforts from specialists in 3D modeling and physics simulation. This process is time-consuming, labor-intensive and limits the scale and variety of training data, becoming a critical barrier to model generalization \cite{jain2025polarisscalablerealtosimevaluations}\cite{doi:10.1177/02783649251396980}\cite{li2024evaluatingrealworldrobotmanipulation}. Second, while automated or procedural generation can create varied scenes, it often lacks fine-grained control. A core tension exists between generating sufficient diversity for robust learning and maintaining the ability to precisely reproduce or systematically vary specific scenarios. This makes it difficult to debug failures, conduct controlled ablation studies, or measure systematic generalization—hindering rigorous model development \cite{guo2025ctrl}. Third, current model evaluation relying on a fixed set of handcrafted metrics (e.g., success rate) fails to capture nuanced task completion quality\cite{geng2025roboverse}. Furthermore, human-in-the-loop evaluation is inefficient, subjective, and non-scalable, resulting in an incomplete assessment of policy performance and slows down the iterative training-evaluation cycle.

To resolve above-mentioned issues, we introduce \textbf{Genie Sim 3.0}, a comprehensive open-source simulation platform for environment reconstruction, scene generalization, data collection and automated evaluation. Genie Sim 3.0 integrates the following four novel features: (1) large language model (LLM)-driven scene generation and generalization; (2) LLM-based task and evaluation generation; (3) high fidelity simulation via 3D reconstruction and visual generative synthesis; (4) dual-mode data collection pipeline, including teleoperation and automation.

Developing on these innovative new features, Genie Sim 3.0 provides five new simulation modules to significantly accelerate the model development of embodied intelligence: (1) Genie Sim asset retrieval system, supporting sematic query-based search for 5,140 simulation-ready objects across 353 categories; (2) a scene generation pipeline that operates solely on natural language input and is equipped with comprehensive domain randomization capabilities; (3) more than 10,000 hours of synthetic data with multi-dimensional domain randomization across 200 tasks; (4) a comprehensive benchmark integrating close-loop evaluation with over 100,000 evaluation scenarios, designed to assess a holistic capability profile across multiple dimensions, covering semantic understanding, spatial reasoning, and operational execution, therefore clearly defining model's performance boundaries and optimization directions. 

Contributions of Genie Sim 3.0 can be summarized into five key aspects: 
\begin{enumerate}
    \item We develop a scene generation pipeline called Genie Sim Generator that features a natural language interface that interprets high-level semantic instructions to generate highly realistic simulation scenes in real-time. This interface supports multi-round conservation for iterative refinement, significantly improving efficiency of scene generation.
    \item We support rapid, multi-dimensional scene generalization from a single generated scene mentioned above. It can produce vast and varied scenarios by parameterizing factors such as lighting, background, layout, poses, trajectories, sensor noise, and robot morphology within several minutes via natural language instructions.
    \item We establish a scalable evaluation benchmark built upon more than 100,000 simulation scenarios. It leverages LLM to automatically generate task instructions and evaluation protocols, which are then executed and scored by VLM. This constructs a comprehensive and multi-dimensional capability profile—spanning semantic understanding, spatial reasoning, and operational execution—that clearly delineates a model’s performance boundaries and optimization pathways.
    \item Through systematic experiments, we validate the effective sim-to-real transfer capability of the proposed benchmark framework and synthetic data. Empirical results demonstrate that, under certain conditions, the generated synthetic data can serve as a viable substitute for real-world robotic data, thereby providing scalable support for model training. Furthermore, the benchmark enables the effective prediction and extrapolation of a model’s performance boundaries in real-world scenarios based on its performance in simulation, offering stable and reproducible guidance for model iteration.
    \item We open-source Genie Sim 3.0 simulation platform which integrates scene generation, evaluation synthesis, data acquisition, and closed-loop model assessment. To foster community progress, we open-source 5,140 assets, 10,000+ hours dataset, 100,000+ evaluation scenarios, and the complete codebase. By providing massive-scale synthetic dataset and evaluation scenes for insightful capability diagnostics and panoramic evaluation to accelerate model iteration, Genie Sim 3.0 reduces reliance on physical hardware, enhances research and development efficiency, and propels the development of practical embodied AI applications.
\end{enumerate}
\section{RELATED WORK}

\subsection{Robotics Dataset}
Numerous open-source datasets are available in robotics, encompassing data collected from both physical robotic platforms and simluation environments. \textit{DROID}~\cite{khazatsky2025droidlargescaleinthewildrobot} dataset provides over 70,000 real-world robotic manipulation demonstrations; \textit{RoboNet}~\cite{dasari2020robonetlargescalemultirobotlearning} integrated 160,000 trajectory data from 7 different robot systems; \textit{RH20T}~\cite{fang2024rh20t} includes 100,000 demonstrations spanning 147 different tasks, reflecting a high degree of task diversity; \textit{Open X-Embodiment}~\cite{embodimentcollaboration2025openxembodimentroboticlearning} integrates vast amounts of heterogeneous data from laboratories worldwide, comprising over 1 million trajectories, 160,266 tasks, and 22 distinct robot embodiments; \textit{Agibot World}~\cite{bu2025agibot_iros} comprises millions of real-robot data trajectories, covering five core scenarios and encompassing over 80 skills. Despite the extensive scale, real-world datasets impose two fundamental limitations: prohibitively high acquisition costs and narrow generalization capability.

Synthetic datasets provide cost-effective, large-scale, diverse data with controlled domain randomization to promote robust policy generalization. \textit{RoboCasa}~\cite{robocasa2024} offers more than 100,000 trajectories, yet its scenes are predominantly confined to kitchen settings; \textit{DexGraspNet 2}~\cite{zhang2024dexgraspnet} contains over 400 million demonstrations but focuses solely on grasping actions without dynamic scene variations or more complex manipulation skills; \textit{RoboTwin 2.0}~\cite{chen2025robotwin} encompasses over 100,000 bimanual manipulation trajectories across 50 tasks, but the physical and visual fidelity of its simulation environment remains notably constrained.

\subsection{Robotics Benchmark}
Benchmarks are essential for assessing model performance, yet creating high-quality benchmarks continues to be a significant challenge with the field. \textit{Meta-World}~\cite{yu2019meta} provides a benchmark designed for multi-task and meta-reinforcement learning. \textit{HumanoidGen}~\cite{jing2025humanoidgen} establishes a dedicated benchmark comprising 20 tabletop manipulation tasks. \textit{HumanoidBench}~\cite{sferrazza2024humanoidbench} contains 27 whole-body control tasks designed to evaluate reinforcement learning and hierarchical control methods. \textit{Bigym}~\cite{chernyadev2024bigym} introduces a benchmark comprising 40 manipulation tasks set in household kitchen scenarios. \textit{BEHAVIOR-1K}~\cite{li2024behavior1k} provides a comprehensive benchmark that includes 50 diverse scenes, over 9,000 object models spanning 1,900+ categories. \textit{ManipulaTHOR}~\cite{ehsani2021manipulathor} is a benchmark consisting of 30 kitchen scenes, over 150 object categories, with defined training, validation and test splits and evaluation metrics. Although seminal benchmarks such as \textit{HomeRobot}~\cite{yenamandra2024homerobotopenvocabularymobilemanipulation}, \textit{DaXBench}~\cite{chen2023daxbench}, \textit{SoftGym}~\cite{corl2020softgym}, \textit{PlasticineLab}~\cite{huang2021plasticinelabsoftbodymanipulationbenchmark}, \textit{INT-ACT}~\cite{fang2025intentionexecutionprobinggeneralization}, \textit{ManiFeel}~\cite{luu2025manifeelbenchmarkingunderstandingvisuotactile}, \textit{AGNOSTOS}~\cite{zhou2025exploring}, \textit{RoboTwin 2}~\cite{chen2025robotwin}, \textit{VLABench}~\cite{zhang2025scenelanguagerepresentingscenes} contribute valuable evaluations, the field remains constrained by several persistent challenges, including narrow evaluation scope, notable sim-to-real gaps in both physical dynamics and visual rendering, and insufficient transfer efficacy to real-world deployments.

% \subsection{Robot Learning}

% Many task-specific policy architectures achieve strong single-task performance but struggle to transfer across embodiments. In contrast, foundation models trained on million-scale, multi-robot corpora have enabled robust zero-shot generalization: RT-1 unifies vision, language and actions in a single transformer for real-time kitchen tasks; RT-2 co-fine-tunes large vision–language models on web and robot data to unlock semantic planning and object reasoning; diffusion-based RDT-1B and the $\pi_0$ capture diverse bimanual dynamics from over a million episodes. Vision–language–action (VLA) frameworks like OpenVLA and CogACT, together with adaptations like Octo, LAPA, and OpenVLA-OFT demonstrate efficient fine-tuning to novel robots and sensor modalities.

\section{Method}

\subsection{Scene Generation}
\begin{figure*}[ht]
    \centering
    \includegraphics[width=1.0\linewidth]{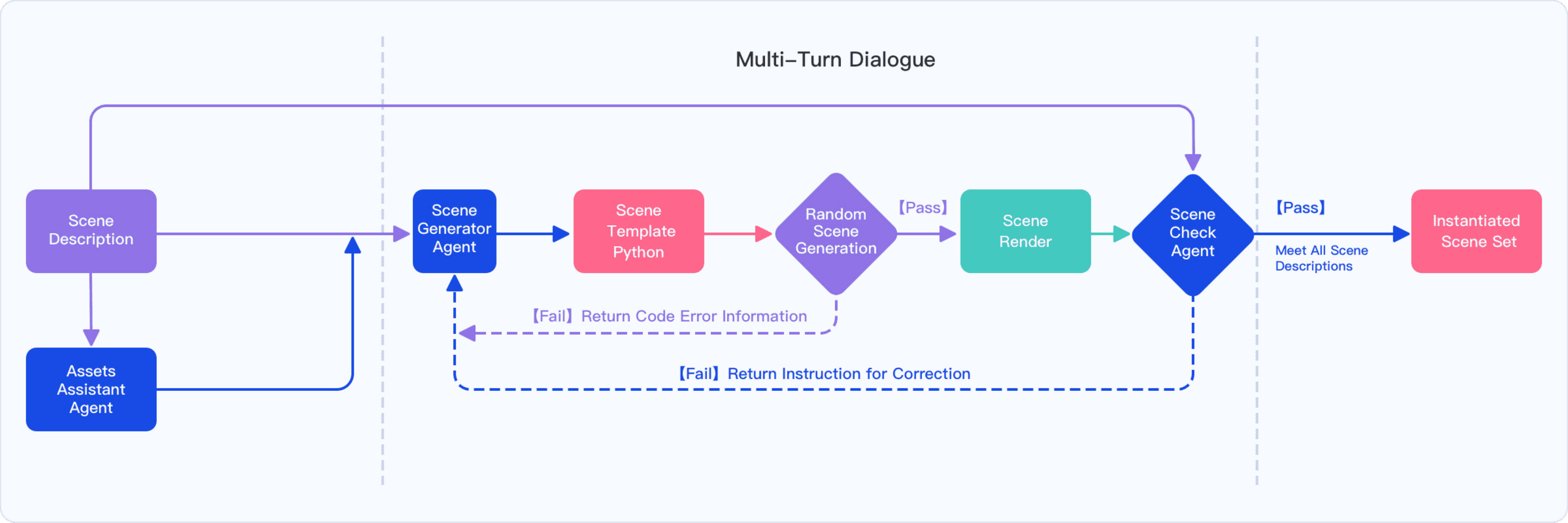}
    \caption{\textbf{The Automated Workflow of Genie Sim Generator}. This module captures user intent via multi-round conversation, translates it into executable Python code, and compiles the final scene graph with assets for Isaac Sim.}
    \label{fig:scene_gen}
\end{figure*}
With the proliferation of LLM and VLM, scalable, generalizable and conversational data creation has become an emerging paradigm.

We implement \textbf{Genie Sim Generator}, an LLM-driven toolchain that jointly creates simulation scenes and produces \textit{Scene Graph} for downstream task evaluations. It consists of two tightly-coupled modules. The first is Assets Index, a managed repository of Isaac Sim-ready assets augmented with LLM, VLM and retrieval-augmented generation (RAG) agents, which offers structured Application Programming Interface (API) to facilitate scene composition. The second is Scene Generator, inspired by the scene language \cite{zhang2025scenelanguagerepresentingscenes}, this module captures user intent via multi-round conversation, translates it into executable Python code, and compiles the final scene graph with assets for Isaac Sim.

The entire scene generation process has four stages: intention interpret, assets retrieve, domain specific language (DSL) code generation and results assemble, all within the same chat window sharing the same context.

\textbf{Intention Interpreter} translates user’s open-ended natural language prompt into a structured, machine-interpretable task request. A chain-of-thought (CoT)–enabled LLM first parses and decomposes the natural language prompt into a structured decomposition comprising spatial scene descriptors, object attribute constraints and task-level intents (e.g., “stack-up”, “tidy”, “random”). Ambiguous or underspecified phrases (e.g., “some blocks”, “randomly arranged”) are resolved through a reasoning process grounded in a pre-trained world-knowledge memory. The resulting specification is formulated as a JavaScript Object Notation (JSON) schema containing required semantic object classes (with optional geometric constraints such as size, color and shape) and pairwise spatial relations (e.g., “on”, “adjacent”, “aligned”). The resulting schema is forwarded to the Assets Index to initiate retrieval process. If constraints contradict predefined system rules, the interpreter engages an explicit feedback loop with the user for clarification.

\textbf{Assets Index} is a RAG-powered asset retrieval module.
Semantic descriptions are first extracted for all 5,140 objects based on their appearance, geometry, and usage. Subsequently, these descriptions are encoded into 2048-dimensional vectors via the QWEN: text-embedding-v4 model and stored in a ChromaDB vector database. At runtime, the planner extracts keywords (e.g., “yellow cube”) from scene description and encodes them into the same embedding space using the identical model. The resulting query vector is compared against the stored asset embeddings via cosine similarity, and the top-k most similar candidates are retrieved along with their metadata—including Universal Scene Description (USD) paths, collision hulls, mass properties, and texture variants. These assets are seamlessly integrated into the LLM context, ensuring that subsequent code generation references only pre-validated, available assets. This retrieval step is fully transparent to the user and typically completes within 200 milliseconds.

\textbf{DSL Code Generater} draws upon the syntactic structure defined in the scene language \cite{zhang2025scenelanguagerepresentingscenes}, but extends it with a back-end adaptation for seamless interoperability with  Genie Sim Assets Library. Integrating the contexts from Intention Interpreter, Assets Index and DSL definition, our system synthesizes a precise scene specification via pre-trained LLM. This representation exhibits fine-grained controllability, double float precision and enhanced generalization capability. We maintain the chat context within our engine to support iterative task editing. The output can also be manually adjusted to address limitations of LLM. By embedding our asset library into the LLM context, we achieve joint generalization over categories, poses, lighting, and textures without any fine-tuning.
Rich semantic annotations and geometric properties within assets library enables scene-level generalization. Complicated layouts which extend beyond basic tabletop scenarios are also supported such as shelves and storage racks.

\textbf{Results Assembler} finally collects the output from chat context and instantiates the DSL program generated by LLM. To achieve randomization, the DSL program employs random functions that introduce variability in object poses, layout patterns, and choice of objects. During instantiation, a hierarchical Scene Graph with two major parts is created: nodes (objectes encoded with asset id, semantic, size, pose and task tag) and edges (spatial relations such as on, in, adjacent, aligned and stacked). Finally assembler utilizes the OpenUSD Schema and Isaac Sim APIs to synthesize simulation-ready USD files. Our Results Assembler generates thousands of diverse scenes within minutes, leveraging the power of LLM and advanced algorithms to produce massive scene layouts with high efficiency.
% \begin{figure}[ht]
%     \centering
%     \includegraphics[width=0.8\linewidth]{figure/instruction_gen.pdf}
%     \caption{LLM-driven tasks generation.}
%     \label{fig:instruction_gen}
% \end{figure}
\begin{figure}[ht]
    \centering
    \includegraphics[width=0.8\linewidth]{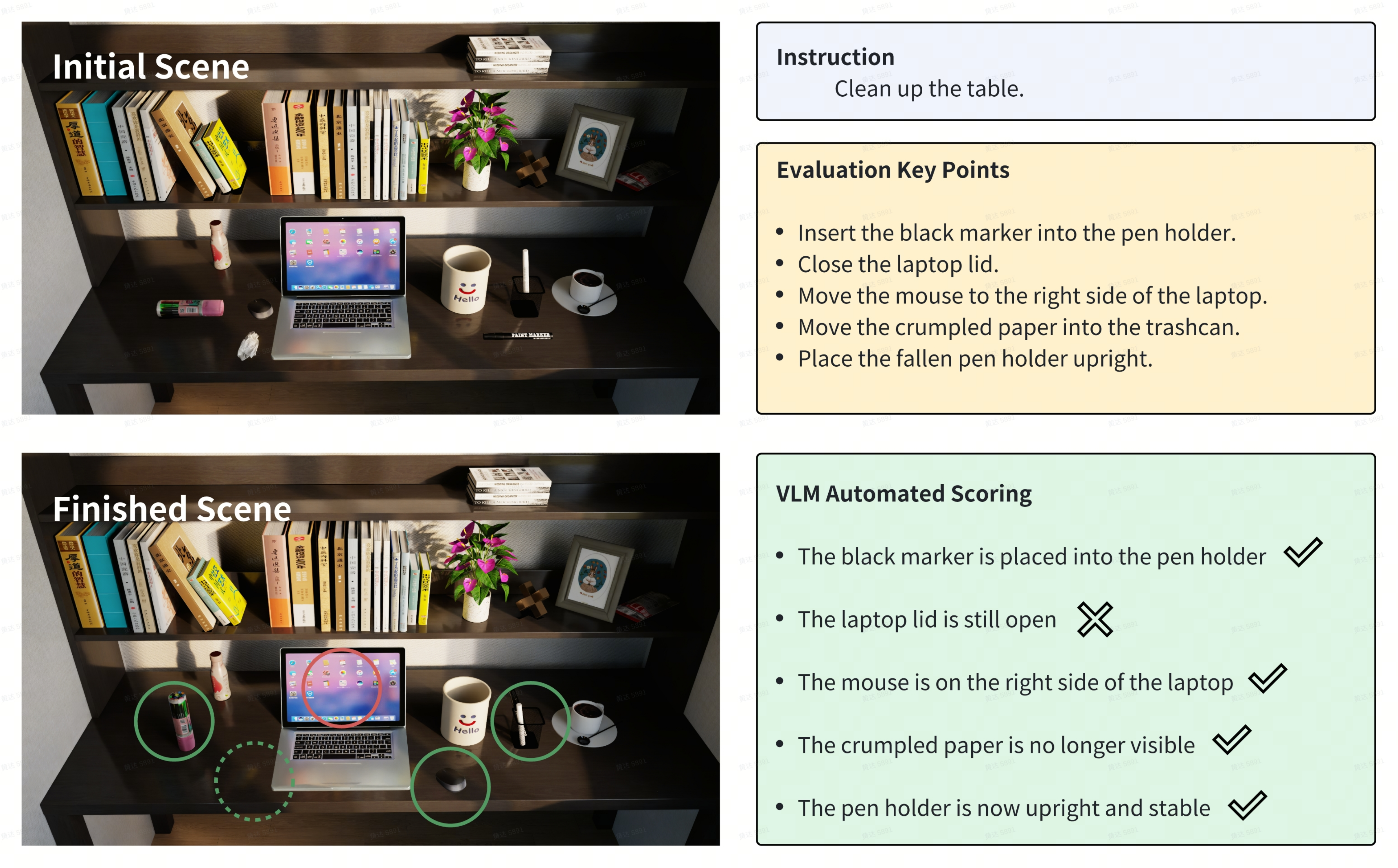}
    \caption{\textbf{VLM-Driven Evaluation}.}
    \label{fig:vlm_autoscore}
\end{figure}
\subsection{Evaluation Generation} 
Current open-source simulation benchmarks typically rely on predefined instructions, manually annotated success criteria and repeated trial executions to evaluate VLA models. This paradigm yields a largely unidimensional instruction space, limits the scalability of evaluations, and entails high costs for new task creation. LLM exhibits strong natural language understanding and therefore can be introduced for automated instruction generation, task decomposition and prompt expansion, enabling more diverse, scalable, and cost‑effective evaluation pipelines.

Given a certain simulation scenario, Genie Sim Benchmark leverages the capabilities of the LLM by combining it with Action Domain Evaluation Rule (ADER) system to automatically generate numerous reasonable instructions and executable evaluation configuration files, which are then integrated with the simulator for automated evaluation.

VLM demonstrates strong visual and semantic understanding. Given formalized task specifications alongside the temporal sequence of visual observations recorded during task execution, VLM can efficiently determine whether task requirements have been satisfied and generate evidence‑based justifications (see Fig. \ref{fig:vlm_autoscore}) . This evaluation paradigm is applicable to the majority of simulation benchmark tasks, markedly reducing human annotation effort and allowing the creation of large‑scale evaluation instances at low cost.

\begin{figure*}[hb]
    \centering
    \includegraphics[width=0.9\linewidth]{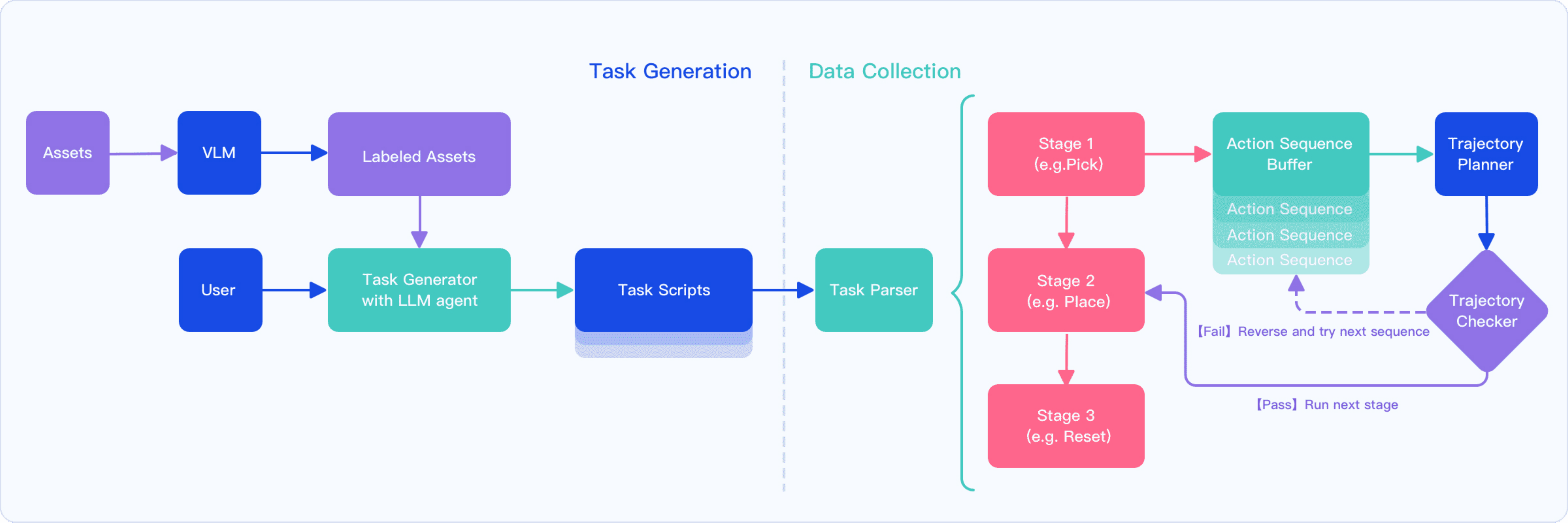}
    \caption{\textbf{Automated Data Collection}. A complete task parsing and execution pipeline, which improves task success rate through waypoint filtering and a robust retry mechanism.}
    \label{fig:data_collection}
\end{figure*}
\subsection{Environment Reconstruction}
To provide large-scale, high-fidelity interactive simulation environment, we utilize 3DGS \cite{kerbl20233d} neural rendering technology for photorealistic image rendering and surface reconstruction, generating high-precision meshes for interactive physical simulation. We collect data using SkylandX Innovation's MetaCam handheld 3D laser scanner. The collected data contains fisheye images, the corresponding poses for each frame, and the dense point cloud data of the entire scene.

However, 3D Gaussian Splatting (3DGS) imposes relatively stringent requirements on camera pose accuracy; deviations of even a few pixels can result in blurred rendering, artifacts, and geometric bulges. In complex indoor environments, LiDAR‑based SLAM often cannot achieve pixel‑level positioning precision, leading to insufficient pose accuracy for high‑quality 3DGS reconstruction.
In the camera pose optimization module, we first utilize SuperPoint \cite{detone2018superpoint} and LightGLue \cite{lindenberger2023lightglue} to replace the DSP-SIFT \cite{dong2015domain} and feature extraction module in COLMAP-PCD \cite{bai2024colmap}, which reduces noise in 2D feature points and enhances the ability to extract features in weak textures.
Then, we perform triangulation directly using the prior poses of the camera obtained by LiDAR SLAM, and search the 2D feature points corresponding to the lidar 3D points for association and the sparse 3D points to implement BA optimization together.
After obtaining the camera pose and 3D point data output using COLMAP-PCD, we train 3DGS using the open-source gsplat \cite{ye2025gsplat}framework.
During the process of large-scale scene reconstruction, it is always inevitable to collect sufficient coverage of the views. To compensate for the insufficient number of views during the collection process and improve the rendering quality, we utilize generative models to extrapolate sufficient  views. We use the pretrained model from Difix3D+ \cite{wu2025difix3d+} to render extrapolated views and obtain high-quality images.
Finally, we conducte 3DGS training based on surface reconstruction using a large number of new perspective images produced by the diffusion model and the corresponding pose data, as well as the lidar point cloud data obtained during data collection. We use PGSR \cite{chen2024pgsr} for surface reconstruction and obtained a high-precision mesh.

% \begin{figure}
%     \centering
%     \includegraphics[width=0.9\linewidth]{figure/scene_reconstruction_framework.pdf}
%     \caption{Scene Reconstruction Framework}
%     \label{fig:task_distribution}
% \end{figure}

\subsection{Data Generation}
The data collection framework integrates two complementary paradigms: teleoperation and automation. Teleoperation is utilized for complex long-horizon tasks to produce high quality human-like demonstrations. In contrast, automated collection excels in efficiency and cost-effectiveness, facilitating the rapid generation of extensive synthetic data for robot imitation learning.

\textbf{Teleoperation:} Our teleoperation framework utilizes a PICO VR Head-Mounted Display (PICO) device to bridge human input with a simulated robotic environment. Functioning as the primary input apparatus, the PICO sends action signals representing the target end-effector pose to a central host. A benchmark module processes these signals, and a motion controller executes the planned trajectory, driving the virtual motion of the robot in simulation environment.

The simulation system incorporates physical effects in real world, including collisions and friction, which enables the robot to manipulate and interactive with simulated objects. The entire interaction sequence, encompassing robot joint states, visual observations, and object poses, is fully logged to constitute the synthetic dataset. Consequently, this approach generates anthropomorphic motion data and leverages human expertise to tackle intricate tasks that are difficult to automate.

\textbf{Automated Collection:} Leveraging \textit{cuRobo}~\cite{sundaralingam2023curobo}—a GPU-accelerated motion planner—as the core trajectory planning module, the fully automated data collection pipeline consists of two parts: task generation and data collection (Figure \ref{fig:data_collection}). Task generation is facilitated by an LLM-based asset retrieval system, through which required simulation assets can be efficiently queried and assembled using predefined atomic skills.

During task parsing, candidate key waypoints—derived from pre-annotated asset configurations, such as grasping poses labeled by \textit{GraspNet}~\cite{DBLP:journals/corr/abs-1912-13470}—are evaluated based on kinematic reachability, collision avoidance, and anthropomorphic feasibility. To improve robustness, multiple candidate waypoints are generated per action, forming several alternative action sequences. Each sequence is executed in simulation and subsequently assessed by a dedicated trajectory evaluation module. In case of execution failure, a state rollback is performed before the next candidate sequence is attempted.

Unlike most previous approaches that simplify the planning environment—for example, objects not directly related to the task are excluded, which can lead to collisions in cluttered scenes—our system retains environmental completeness. In order to balance completeness and efficiency, mesh simplification is applied to object geometries during scene initialization, maintaining trajectory reliability while significantly improving computational efficiency during planning.

\subsection{Closed-loop Evaluation}

For model closed-loop evaluation, the simulation and the model inference environment are decoupled and communicate over the Hypertext Transfer Protocol (HTTP) protocol. The simulator transmits the robot’s observation images and proprioceptive states to the inference service and the model returns control commands, which are then executed in the simulation environment. During simulation execution, task completion is periodically evaluated; if the task is completed, the simulation terminates, otherwise a predefined timeout procedure is invoked to terminate the run.
Genie Sim 3.0 benchmark supports the following evaluation features:
\begin{itemize}
\item Integration with common VLA models (\textit{$\pi_{0.5}$}~\cite{intelligence2025pi05visionlanguageactionmodelopenworld}, \textit{GO-1}~\cite{bu2025agibot_iros}, GR00T, \textit{UniVLA}~\cite{bu2025univla}, \textit{RDT}~\cite{liu2024rdt}, \textit{X-VLA}~\cite{zheng2025x}, etc.)
\item Multiple robot types (Genie G1, Genie G2)
\item Multiple end effectors (omnipicker, omnihands, INSPIRE skillhands, zhixing gripper, etc.)
\item Local/distributed inference framework
\item Automated and multi-dimensional evaluation
\end{itemize}

\section{Dataset}

\subsection{Task Distribution}
To accommodate developers at different stages who may prioritize varying levels of data complexity and focus, our dataset is designed to provide comprehensive coverage of the core competencies required by contemporary VLA models for embodied intelligence. The complexity of the tasks is hierarchically organized, facilitating a structured progression from simple to complex scenarios. A key design principle is composability: long-horizon tasks can be decomposed into sequences of fundamental sub-tasks contained within the dataset.

\begin{figure}
    \centering
    \includegraphics[width=0.8\linewidth]{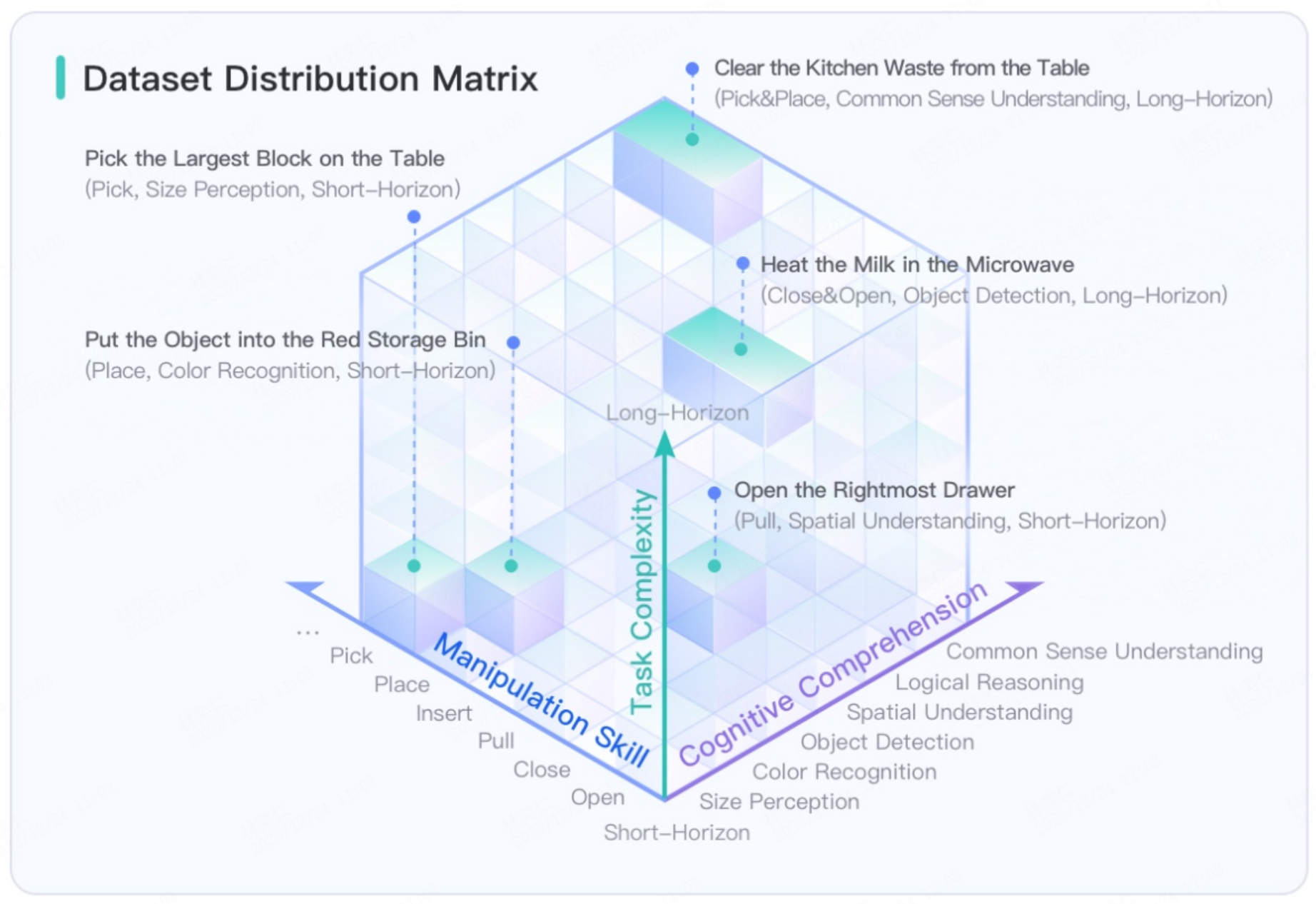}
    \caption{\textbf{Task Distribution Matrix}. The dataset is constructed along three dimensions: Manipulation skill, Cognitive comprehension, and Task complexity.}
    \label{fig:task_distribution}
\end{figure}

To this end, we structure the task taxonomy along three primary axes (Figure \ref{fig:task_distribution}): manipulation skill, cognitive comprehension, and task complexity.
\begin{itemize}
    \item \textbf{Atomic Skills} encompass the model's fundamental motor actions, such as \textit{pick}, \textit{place}, \textit{pull}, \textit{push}, \textit{open}, and \textit{close}.
    \item \textbf{Cognitive Comprehension} pertain to the model's ability to interpret tasks, scenes, and instructions. This includes competencies like \textit{spatial reasoning} (e.g., size relationships), \textit{attribute understanding} (e.g., color), \textit{logical inference}, and \textit{commonsense reasoning}.
    \item \textbf{Task Complexity} is defined along metrics such as planning horizon and the need for coordinated control. For instance, the sequence from ``single-arm removal of a trash item'' to ``bimanual coordinated removal of a trash item'' and finally to ``cleaning all trash from a desktop'' illustrates a graduated increase in complexity.
\end{itemize}
\subsection{Data Distribution}

Building upon the aforementioned principles of task distribution, we introduce Genie Sim 3.0 synthetic dataset. This dataset encompasses 200 representative tasks for embodied intelligence, providing over 10,000 hours of simulated interaction that reflects its substantial scale. The data is generated using two robotic platforms, Agibot's G1 and G2, and incorporates systematic variations across multiple dimensions—including task layout, initial robot pose, environmental lighting, scene configuration, camera noise, and semantic instruction phrasing—ensuring broad coverage and strong generalization capability. The resulting dataset thus supports the training and evaluation of embodied intelligence models with high-quality, diverse, and robust simulated experiences.

\section{Experiment}

To systematically assess the validity and utility of GenieSim as a simulation benchmark and data generation platform, we introduce five closed-loop evaluation suites. \textbf{GenieSim-Sim2Real} is designed to validate the data quality of GenieSim and the fidelity of its simulated environments: it examines whether large-scale synthetic data can match or exceed real-world data in model training, and whether benchmark evaluations in simulation yield conclusions consistent with physical deployment. \textbf{GenieSim-Instruction} focuses on language-conditioned manipulation, probing how well policies can follow diverse and compositional natural language instructions. \textbf{GenieSim-Robust} systematically evaluates policy robustness by introducing controlled perturbations spanning language instruction rephrasing, robot initial pose, scene background and lighting, camera image quality, and camera position. \textbf{GenieSim-Manipulation} covers a broad set of challenging manipulation tasks that require precise motor skills, long-horizon planning, and dexterous object interactions. \textbf{GenieSim-Spatial} probes spatial understanding and reasoning, requiring policies to resolve absolute and relative spatial references, perform relational placement, order objects by attribute, and construct vertical structures. Together, these five suites provide complementary perspectives on policy capabilities and benchmark reliability.

\begin{figure}[ht]
    \centering
    \includegraphics[width=1.0\linewidth]{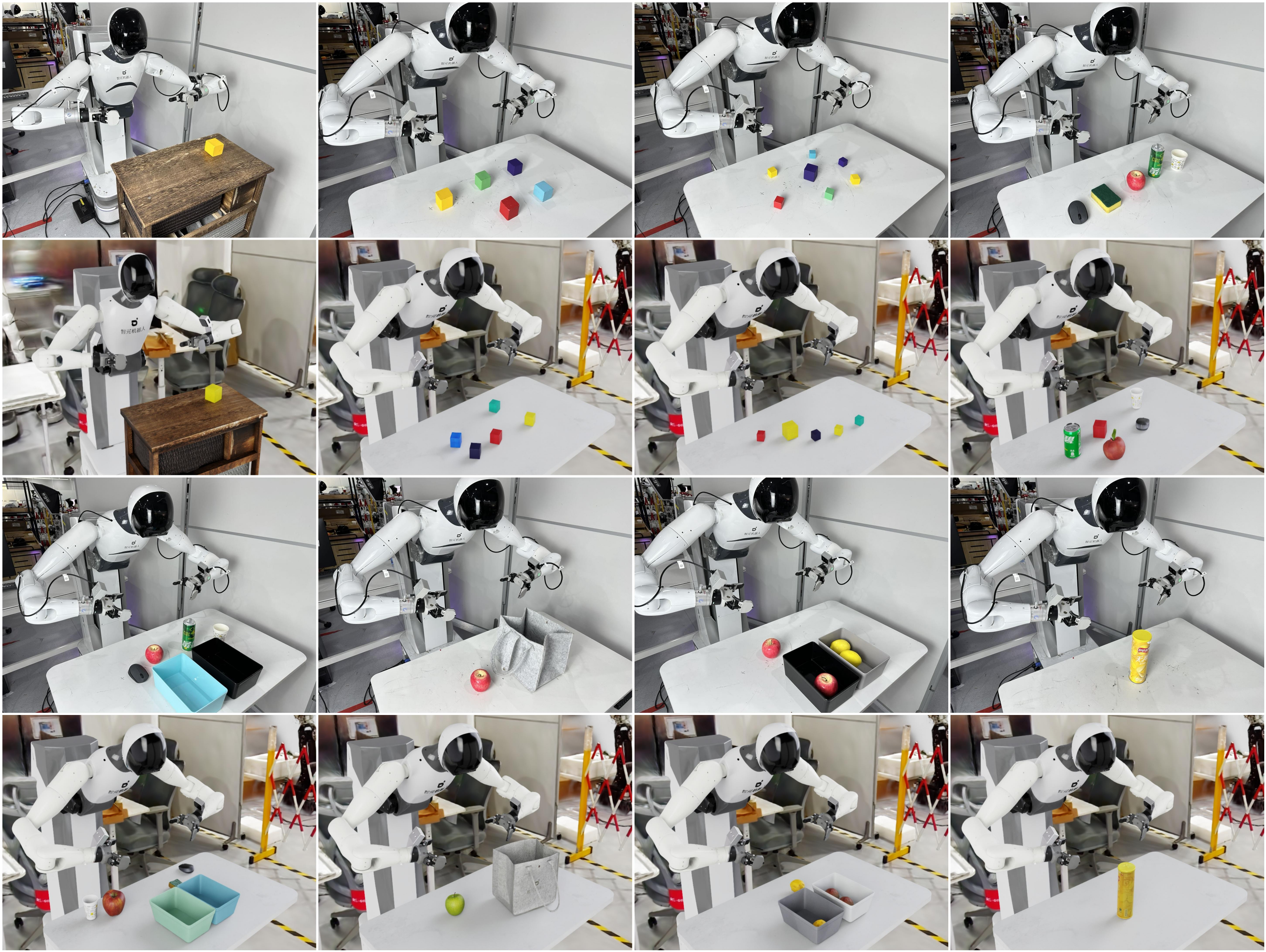}
    \caption{\textbf{GenieSim-Sim2Real Task Setups}. Comparison of initial task configurations between real and simulated testing environments across all evaluated tasks.}
    \label{fig:experiments}
\end{figure}

\subsection{Experimental Setup}

\noindent\textbf{GenieSim-Sim2Real.}
All experiments within this suite use the $\pi_{0.5}$ model~\cite{intelligence2025pi05visionlanguageactionmodelopenworld} as the base policy, post-trained on varying amounts of real-world or synthetic data. The Agibot G1 robot is deployed for physical task execution. Each training configuration is evaluated over 50 trials in real-world environments and in simulation, ensuring direct comparability between the two evaluation modalities.

\noindent\textbf{GenieSim-Instruction, GenieSim-Robust, GenieSim-Manipulation, and GenieSim-Spatial.}
For these four benchmark suites, four state-of-the-art vision-language-action models are compared: $\pi_{0.5}$~\cite{intelligence2025pi05visionlanguageactionmodelopenworld}, ACoT-VLA~\cite{zhong2026acotvlaactionchainofthoughtvisionlanguageaction}, GR00T-N1.7~\cite{gr00tn1_2025}, and $\pi_0$ ~\cite{black2026pi0visionlanguageactionflowmodel}. Each model is fine-tuned with task-specific synthetic data generated by GenieSim and evaluated on the Agibot G2 robot using GenieSim's automated closed-loop evaluation pipeline.

\subsection{GenieSim-Sim2Real}
\label{sec:sim2real}

\subsubsection{Synthetic Data Scaling}
\label{sec:scaling}

\setlength{\tabcolsep}{12pt}
\begin{table*}[ht]
    \caption{\textbf{Synthetic Data Scaling Results.} Task scores of $\pi_{0.5}$ fine-tuned with varying amounts of real and synthetic data, evaluated in both simulated and real environments.}
    \label{table:1}
    \centering
    \begin{tabular}{c cc cc cc cc}
        \toprule
        & \multicolumn{2}{c}{Select Color} & \multicolumn{2}{c}{Recognize Size} & \multicolumn{2}{c}{Grasp Targets} & \multicolumn{2}{c}{Organize Objects} \\
        & sim env  & real env & sim env & real env & sim env & real env & sim env & real env \\
        \midrule
        200 eps real & 0.45 & 0.53 & 0.50 & 0.56 & 0.34 & 0.39 & 0.25 & 0.30 \\
        500 eps real & 0.75 & 0.73 & 0.75 & 0.75 & 0.54 & 0.58 & 0.45 & 0.40 \\
        500 eps sim & 0.53 & 0.60 & 0.50 & 0.63 & 0.29 & 0.33 & 0.39 & 0.35 \\
        1500 eps sim & \textbf{0.86} & \textbf{0.85} & \textbf{0.93} & \textbf{0.94} & \textbf{0.72} & \textbf{0.71} & \textbf{0.48} & \textbf{0.60} \\
        \bottomrule
    \end{tabular}
\end{table*}

We conduct 32 groups of experiments across four representative tasks: Select Color, Recognize Size, Grasp Targets, and Organize Items. These tasks span a range of complexities---from atomic manipulation skills to language-conditioned cognitive reasoning---while avoiding floor or ceiling effects that would obscure performance differences. For each task, $\pi_{0.5}$ is fine-tuned under four data configurations: 200 and 500 episodes of real-world data, and 500 and 1500 episodes of synthetic data. Results are reported in Tab.~\ref{table:1}.

\noindent\textbf{Effectiveness of Synthetic Data.}
Scores improve monotonically with increasing data volume for both real and synthetic sources, consistent with scaling laws in robot learning. At equivalent scale (500 eps), models trained on real-world data outperform those trained on synthetic data, reflecting the higher physical fidelity of real demonstrations. However, scaling synthetic data to 1500 episodes enables the model to surpass all real-data baselines across all four tasks in zero-shot real-world evaluation. This demonstrates that GenieSim's systematic domain randomization---spanning object textures, lighting conditions, physics parameters, and task variations---effectively bridges the domain gap at scale, offering a practical and scalable alternative to real-world data collection.

\begin{figure}[ht]
    \centering
    \includegraphics[width=0.8\linewidth]{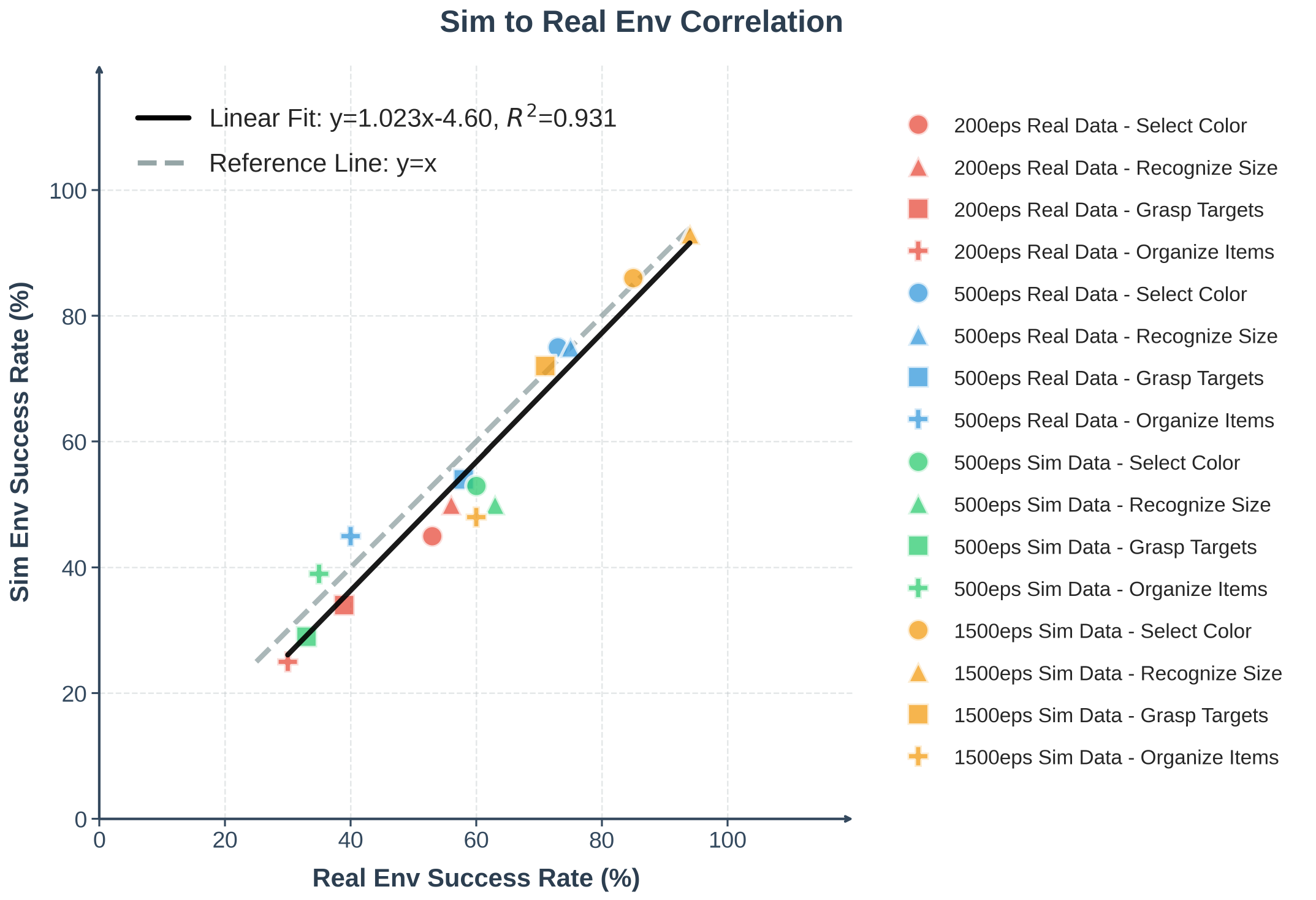}
    \caption{\textbf{Correlation Analysis of Sim and Real Performance}. All 16 model configurations show a strong linear relationship between simulated and real-world scores ($R^2 = 0.931$).}
    \label{fig:sim_real_scatter}
\end{figure}

\noindent\textbf{Consistency between Sim and Real Environments.}
Model performance across simulated and physical environments is compared. Despite minor discrepancies attributable to imperfect simulation of contact dynamics and stochastic real-world perturbations, overall performance trends remain consistent across environments. A quantitative correlation analysis (Fig.~\ref{fig:sim_real_scatter}) yields $R^2 = 0.931$ with a slope $\approx 1.023$, indicating that GenieSim evaluations reliably predict real-world performance trends and can serve as a cost-efficient substitute for physical benchmarking.

\subsubsection{Sim-to-Real Transfer}

\begin{table*}[t]
 \centering
 \caption{\textbf{Sim-to-Real Transfer Results.} $\pi_{0.5}$ scores for models trained on synthetic or real data, evaluated in both simulated and real environments across 8 tasks.\textsuperscript{$\dagger$}}
 \label{tab:sim_real_comparison}
 \begin{tabular}{ccccc}
 \toprule
 \multirow{3.5}{*}{\textbf{Tasks}} & \multicolumn{2}{c}{Sim Env} & \multicolumn{2}{c}{Real Env} \\
 \cmidrule(lr){2-3} \cmidrule(lr){4-5}
  & \textit{Sim Data} & \textit{Real Data} & \textit{Sim Data} & \textit{Real Data} \\
  & (sim-to-sim) & (real-to-sim) & (sim-to-real) & (real-to-real) \\
 \midrule
 Select Color             & \textbf{0.86} & 0.75 & \textbf{0.85} & 0.73 \\
 Recognize Size           & \textbf{0.93} & 0.75 & \textbf{0.94} & 0.75 \\
 Grasp Targets            & \textbf{0.72} & 0.54 & \textbf{0.71} & 0.58 \\
 Organize Items           & \textbf{0.48} & 0.45 & \textbf{0.60} & 0.40 \\
 Pack in Supermarket      & 0.94 & \textbf{1.00} & \textbf{0.95} & \textbf{0.95} \\
 Sort Fruit               & \textbf{0.90} & \textbf{0.90} & \textbf{1.00} & \textbf{1.00} \\
 Place Block into Drawer  & 0.80 & \textbf{0.90} & 0.85 & \textbf{0.90} \\
 Bimanual Chip Handover   & \textbf{0.80} & 0.70 & \textbf{0.73} & 0.71 \\
 \midrule
 \textbf{Avg.}            & \textbf{0.80} & 0.75 & \textbf{0.83} & 0.75 \\
 \bottomrule
 \end{tabular}
 \vspace{2pt}

 \centering
 \scriptsize\textit{\textsuperscript{$\dagger$} Sim Data: $500 \sim 1500$ episodes of simulation data. Real Data: $500$ episodes of real-world data. All models are post-trained from the $\pi_{0.5}$ baseline.}
 \end{table*}

To further validate the cross-domain transfer capability of GenieSim, we extend the evaluation to 8 tasks in Tab.~\ref{tab:sim_real_comparison}, adding four newly introduced tasks---Pack in Supermarket, Sort Fruit, Place Block into Drawer, and Bimanual Chip Handover---to the original four. For each task, models trained on synthetic data ($500$--$1500$ episodes) are compared against models trained on 500 episodes of real-world data, with both evaluated in simulated and physical environments.

\noindent\textbf{Sim-to-Real.}
Across all 8 tasks, models trained solely on synthetic data achieve competitive or superior real-world performance relative to models trained on real data, with an average sim-to-real score of 0.83 versus 0.75 for real-to-real. This confirms that GenieSim-generated data is of sufficient fidelity for direct real-world deployment.

\noindent\textbf{Real-to-Sim.}
Conversely, models trained on real-world data also transfer well to simulated evaluation, achieving an average of 0.75. This bidirectional transferability demonstrates that the GenieSim environment faithfully captures the physical structures and task dynamics of real-world settings, establishing it as a reliable evaluation platform.

Taken together, the 32-group scaling experiments and the 8-task transfer study demonstrate that GenieSim provides both high-quality synthetic data and a reliable simulation benchmark---the two foundational properties required for the credible cross-model evaluations conducted in the following suites.

\subsection{GenieSim Benchmark Evaluation}
\label{sec:benchmark}

Grounded in the validated reliability of GenieSim, we present a systematic comparison of $\pi_{0.5}$, ACoT-VLA, GR00T-N1.7, and $\pi_0$ across four complementary benchmark suites. Each suite targets a distinct capability dimension---instruction following, perturbation robustness, manipulation dexterity, and spatial reasoning---and the high sim-to-real consistency established in Sec.~\ref{sec:sim2real} ensures that simulation-based rankings are predictive of real-world policy performance.

\noindent\textbf{GenieSim-Instruction.}
This suite comprises 10 tasks that require policies to interpret and execute diverse natural language instructions, including directives based on object number, shape, type, color, size, and commonsense or logical reasoning. The tasks are designed to stress-test the language grounding capabilities of VLA models beyond simple keyword matching. As shown in Tab.~\ref{tab:instruction_results}, ACoT-VLA and $\pi_{0.5}$ attain the top average scores of 0.76 and 0.75, substantially outperforming GR00T-N1.7 (0.65) and $\pi_0$ (0.37). The two leaders are separated by only 0.01 on average yet differ systematically by task type: ACoT-VLA leads on perception-intensive attribute-matching tasks such as pick\_block\_color (0.96), pick\_billiards\_color (0.91), pick\_specific\_object (0.83), and pick\_block\_size (0.83), whereas $\pi_{0.5}$ holds an edge on tasks requiring counting, geometric abstraction, or symbolic composition---pick\_block\_number (0.82), pick\_block\_shape (0.55), pick\_follow\_logic\_or (0.81), and straighten\_object (0.59). GR00T-N1.7 tops the field on pick\_common\_sense (0.68), the single task that most depends on world knowledge rather than on visual attribute grounding. Two tasks remain difficult for every model: pick\_block\_shape (0.44--0.55 for the three strong policies) requires abstracting geometry from appearance, and pick\_common\_sense (0.59--0.68) requires knowledge that cannot be read off the scene. The consistently low scores of $\pi_0$ (0.19 on pick\_common\_sense, 0.20 on pick\_block\_shape) confirm that a stronger language-model backbone provides a decisive advantage in instruction-following scenarios.

\begin{table*}[t]
 \centering
 \caption{\textbf{Experimental results on GenieSim-Instruction.} Scores across 10 language-conditioned manipulation tasks.}
 \label{tab:instruction_results}
 \begin{tabular}{lcccc}
 \toprule
 \textbf{Tasks} & $\boldsymbol{\pi_{0.5}}$ & ACoT-VLA & GR00T-N1.7 & $\boldsymbol{\pi_0}$ \\
 \midrule
 pick\_billiards\_color & 0.83 & \textbf{0.91} & 0.73 & 0.50 \\
 pick\_block\_color & 0.87 & \textbf{0.96} & 0.80 & 0.49 \\
 pick\_block\_number & \textbf{0.82} & 0.76 & 0.61 & 0.40 \\
 pick\_block\_shape & \textbf{0.55} & 0.52 & 0.44 & 0.20 \\
 pick\_block\_size & 0.79 & \textbf{0.83} & 0.62 & 0.34 \\
 pick\_common\_sense & 0.59 & 0.66 & \textbf{0.68} & 0.19 \\
 pick\_follow\_logic\_or & \textbf{0.81} & 0.79 & 0.72 & 0.39 \\
 pick\_object\_type & \textbf{0.80} & \textbf{0.80} & 0.72 & 0.42 \\
 pick\_specific\_object & 0.81 & \textbf{0.83} & 0.69 & 0.37 \\
 straighten\_object & \textbf{0.59} & 0.51 & 0.45 & 0.38 \\
 \midrule
 \textbf{Avg.} & 0.75 & \textbf{0.76} & 0.65 & 0.37 \\
 \bottomrule
 \end{tabular}
 \end{table*}

\begin{table*}[ht]
  \centering
  \caption{\textbf{Experimental results on GenieSim-Robust.} Task scores under 5 categories of perturbations applied to a reference pick-and-place task, anchored by the unperturbed Reference score. $\Delta$ is the signed change relative to Reference, where a negative value marks degradation and a positive value an improvement.}
  \label{tab:robustness_results}
  \begin{tabular}{l cc cc cc cc}
  \toprule
  & \multicolumn{2}{c}{$\boldsymbol{\pi_{0.5}}$} & \multicolumn{2}{c}{ACoT-VLA} & \multicolumn{2}{c}{GR00T-N1.7} & \multicolumn{2}{c}{$\boldsymbol{\pi_0}$} \\
  \cmidrule(lr){2-3}\cmidrule(lr){4-5}\cmidrule(lr){6-7}\cmidrule(lr){8-9}
  \textbf{Perturbation} & Score & $\Delta$ & Score & $\Delta$ & Score & $\Delta$ & Score & $\Delta$ \\
  \midrule
  Reference        & 0.750 & --       & \textbf{0.760} & --       & 0.650 & --       & 0.370 & --       \\
  Instruction      & 0.689 & $-0.061$ & \textbf{0.739} & $-0.021$ & 0.678 & $+0.028$ & 0.332 & $-0.038$ \\
  Robot Pose       & 0.492 & $-0.258$ & \textbf{0.543} & $-0.217$ & 0.429 & $-0.221$ & 0.240 & $-0.130$ \\
  Background       & 0.709 & $-0.041$ & \textbf{0.737} & $-0.023$ & 0.646 & $-0.004$ & 0.381 & $+0.011$ \\
  Image Quality    & \textbf{0.599} & $-0.151$ & 0.589 & $-0.171$ & 0.489 & $-0.161$ & 0.307 & $-0.063$ \\
  Camera Position  & \textbf{0.575} & $-0.175$ & 0.504 & $-0.256$ & 0.449 & $-0.201$ & 0.303 & $-0.067$ \\
  \midrule
  \textbf{Avg.}    & 0.613 & $-0.137$ & \textbf{0.622} & $-0.138$ & 0.538 & $-0.112$ & 0.313 & $-0.057$ \\
  \bottomrule
  \end{tabular}
\end{table*}

\begin{table*}[ht]
  \centering
  \caption{\textbf{Experimental results on GenieSim-Manipulation.} Scores across 10 complex manipulation tasks.}
  \label{tab:manipulation_results}
  \begin{tabular}{lcccc}
  \toprule
  \textbf{Tasks} & $\boldsymbol{\pi_{0.5}}$ & ACoT-VLA & GR00T-N1.7 & $\boldsymbol{\pi_0}$ \\
  \midrule
  Open Door                          & \textbf{0.95} & 0.55 & 0.90 & 0.90 \\
  Hold Pot                           & \textbf{0.55} & 0.45 & 0.11 & 0.20 \\
  Pour Workpiece                     & \textbf{0.93} & \textbf{0.93} & 0.87 & 0.07 \\
  Stock and Straighten Shelf         & \textbf{0.40} & 0.18 & 0.24 & 0.18 \\
  Take Wrong Item Shelf              & \textbf{0.85} & \textbf{0.85} & 0.77 & 0.65 \\
  Scoop Popcorn                      & \textbf{0.95} & \textbf{0.95} & 0.77 & 0.87 \\
  Clean the Desktop                  & 0.00 & 0.00 & \textbf{0.02} & 0.00 \\
  Place Block into Box               & \textbf{0.58} & \textbf{0.58} & 0.50 & 0.45 \\
  Sorting Packages                   & \textbf{0.51} & 0.28 & 0.20 & 0.15 \\
  Sorting Packages Continuous        & \textbf{0.10} & 0.00 & 0.00 & 0.00 \\
  \midrule
  \textbf{Avg.}                      & \textbf{0.58} & 0.48 & 0.44 & 0.35 \\
  \bottomrule
  \end{tabular}
  \end{table*}

\begin{table*}[ht]
  \centering
  \caption{\textbf{Experimental results on GenieSim-Spatial.} Scores across 8 spatial-reasoning tasks.}
  \label{tab:spatial_results}
  \begin{tabular}{lcccc}
  \toprule
  \textbf{Tasks} & $\boldsymbol{\pi_{0.5}}$ & ACoT-VLA & GR00T-N1.7 & $\boldsymbol{\pi_0}$ \\
  \midrule
  Pick Object Absolute Position          & 0.80 & \textbf{0.81} & 0.55 & 0.46 \\
  Pick Object Relative Position          & \textbf{0.65} & 0.60 & 0.43 & 0.20 \\
  Place Beverage to Another's Position   & 0.40 & \textbf{0.53} & 0.35 & 0.11 \\
  Place Object Relative Position         & 0.50 & \textbf{0.51} & 0.43 & 0.21 \\
  Sort Cubes by Size                     & 0.00 & \textbf{0.30} & 0.10 & 0.00 \\
  Sort Number                            & 0.00 & \textbf{0.13} & \textbf{0.13} & 0.06 \\
  Stack Bowls                            & \textbf{0.20} & 0.00 & 0.00 & 0.00 \\
  Stack Three Building Blocks            & \textbf{0.30} & \textbf{0.30} & 0.00 & 0.00 \\
  \midrule
  \textbf{Avg.}                          & 0.36 & \textbf{0.40} & 0.25 & 0.13 \\
  \bottomrule
  \end{tabular}
  \end{table*}

\noindent\textbf{GenieSim-Robust.}
This suite systematically evaluates policy robustness by introducing 5 categories of controlled perturbations to a reference pick-and-place task, spanning language instruction rephrasing, robot initial pose, scene background and lighting, camera image quality, and camera position. These conditions model realistic deployment variabilities that are difficult to enumerate exhaustively in physical experiments but are readily programmable in simulation.

As reported in Tab.~\ref{tab:robustness_results}, we anchor each model against its unperturbed Reference score and report the signed change $\Delta$ under every perturbation. ACoT-VLA and $\pi_{0.5}$ are nearly tied at the top (0.622 and 0.613 on average), well ahead of GR00T-N1.7 (0.538) and $\pi_0$ (0.313); ACoT-VLA is strongest under Instruction, Robot Pose, and Background, and $\pi_{0.5}$ under Image Quality and Camera Position. The $\Delta$ columns split sharply by perturbation type: language- and appearance-level shifts are essentially free---Instruction and Background cost at most 0.061 and are even mildly beneficial for GR00T-N1.7 ($+0.028$) and $\pi_0$ ($+0.011$)---whereas geometric and embodiment-level shifts dominate the degradation, with Robot Pose and Camera Position each removing up to $\sim$0.26 and Image Quality remaining intermediate. Crucially, $\Delta$ must be read jointly with the absolute score: $\pi_0$ exhibits the \emph{smallest} average degradation (0.057), yet this reflects a floor effect from its low Reference of 0.370 rather than genuine robustness, since its perturbed scores remain the worst of all models. Strong clean-task competence---not a small relative drop---is the decisive driver of deployable robustness.

\noindent\textbf{GenieSim-Manipulation.}
This suite consists of 10 tasks that target diverse and challenging manipulation skills, including articulated object interaction (Open Door), bimanual collaboration (Hold Pot), granular material handling (Scoop Popcorn), precise placement and sorting (Place Block into Box, Sorting Packages), desktop cleaning, and multi-step shelf organization. These tasks collectively probe the breadth and depth of a policy's physical manipulation capabilities in realistic scenarios. As shown in Tab.~\ref{tab:manipulation_results}, $\pi_{0.5}$ achieves the highest average score of 0.58, ahead of ACoT-VLA (0.48), GR00T-N1.7 (0.44), and $\pi_0$ (0.35). High-performance tasks such as Pour Workpiece, Take Wrong Item Shelf, and Scoop Popcorn are handled competitively by multiple models, reflecting the robustness of large pretrained VLA backbones on well-structured manipulation primitives. In contrast, long-horizon sequential tasks---Sorting Packages Continuous (0.10 / 0.00 / 0.00 / 0.00) and Clean the Desktop (0.00 / 0.00 / 0.02 / 0.00)---remain near-unsolved for all models, highlighting the frontier of current policy capabilities and the discriminative value of this suite.

\noindent\textbf{GenieSim-Spatial.}
This newly introduced suite comprises 8 tasks that probe spatial understanding and reasoning, spanning absolute and relative spatial reference (Pick Object Absolute / Relative Position), relational placement (Place Beverage to Another's Position, Place Object Relative Position), attribute-based ordering (Sort Cubes by Size, Sort Number), and vertical construction (Stack Bowls, Stack Three Building Blocks). As shown in Tab.~\ref{tab:spatial_results}, ACoT-VLA achieves the highest average of 0.40, followed by $\pi_{0.5}$ (0.36), GR00T-N1.7 (0.25), and $\pi_0$ (0.13). The suite separates cleanly into two regimes. \emph{Referential} tasks---resolving a spatial expression to a target or goal pose---are largely tractable, with the two leaders reaching 0.80 and 0.81 on Pick Object Absolute Position. \emph{Constructive} tasks---ordering objects by attribute or stacking them into a stable structure---collapse for every model: no policy exceeds 0.30 on Sort Cubes by Size, Sort Number, Stack Bowls, or Stack Three Building Blocks. The split holds even at the bottom of the ranking, where $\pi_0$ reaches 0.46 on absolute-position grounding yet stays at or below 0.06 on all four constructive tasks. The leaders remain complementary---ACoT-VLA tops both attribute-ordering tasks, while $\pi_{0.5}$ is alone in making progress on Stack Bowls (0.20)---but the 0.38 -- 0.46 mean gap between the two regimes for the three strong policies identifies sequential spatial construction, rather than spatial reference resolution, as the binding constraint on current VLA policies.

Across the four task suites, $\pi_{0.5}$ and ACoT-VLA emerge as the two strongest policies under complementary profiles. ACoT-VLA holds a narrow lead on the three cognition-oriented suites---GenieSim-Instruction (0.76 vs.\ 0.75), GenieSim-Robust (0.622 vs.\ 0.613), and GenieSim-Spatial (0.40 vs.\ 0.36)---while $\pi_{0.5}$ is decisively ahead on GenieSim-Manipulation (0.58 vs.\ 0.48), the suite that most heavily weights long-horizon motor execution over language and spatial grounding. The margins on the first three suites (0.01 -- 0.04) are far smaller than the 0.10 gap on the fourth, indicating that action-chain-of-thought supervision buys reasoning quality rather than manipulation dexterity. GR00T-N1.7 follows as a competitive third policy, staying within reach of the leaders on several suites (0.65 on GenieSim-Instruction and 0.538 on GenieSim-Robust) and topping individual tasks such as pick\_common\_sense (0.68) and Clean the Desktop, while $\pi_0$ trails by a clear margin throughout (0.37 / 0.313 / 0.35 / 0.13). Read jointly, the four suites also localize where the frontier lies: instruction following is largely solved for strong backbones (leaders at or above 0.75), whereas long-horizon sequential execution---Sorting Packages Continuous, Clean the Desktop, and the four spatial construction tasks---remains at or below 0.30 for every model. The credibility of these cross-model comparisons rests on the reliability of GenieSim as a benchmark platform, which has been rigorously validated through the sim-to-real experiments in Sec.~\ref{sec:sim2real}.

\section{CONCLUSIONS}
We present Genie Sim 3.0, a unified platform that addresses key bottlenecks in robot learning through these integrated contributions: Genie Sim Generator that uses LLM to create generalized, high-fidelity scenes from natural language instructions; a pioneering LLM-VLM benchmark for automated, scalable task generation and evaluation; and a large-scale open-source dataset validated for effective zero-shot sim-to-real transfer. Together, these components constitute a cohesive foundation for advancing scalable data generation, robust policy assessment, and reproducible research toward more generalizable robotic manipulation.


%%%%%%%%%%%%%%%%%%%%%%%%%%%%%%%%%%%%%%%%%%%%%%%%%%%%%%%%%%%%%%%%%%%%%%%%%%%%%%%%

%%%%%%%%%%%%%%%%%%%%%%%%%%%%%%%%%%%%%%%%%%%%%%%%%%%%%%%%%%%%%%%%%%%%%%%%%%%%%%%%

%%%%%%%%%%%%%%%%%%%%%%%%%%%%%%%%%%%%%%%%%%%%%%%%%%%%%%%%%%%%%%%%%%%%%%%%%%%%%%%%
% \bibliographystyle{plain}
% \bibliography{reference}

\section*{APPENDIX}

\section*{ACKNOWLEDGMENT}
We would like to express our sincere gratitude to Linqing Zhong, Sukai Wang and Xiaowei Cai for their assistance with model inference and training; to Haowen Yin, Yanping Zhou, Jiahao Yu, Zengcheng Zhou, and Cheng Ruan for their support with experiment testing; to Chengyue Zhao, Xinke Yu, and Jialu Li for their work on publicity and design; and to Cheng Jing, Haoyu Cao, Chi Zhang, Guangte Xiang, Jun Dai, Jia Zeng for their help with data processing.
\printbibliography

@INPROCEEDINGS{Chi-RSS-23, 
    AUTHOR    = {Cheng Chi AND Siyuan Feng AND Yilun Du AND Zhenjia Xu AND Eric Cousineau AND Benjamin CM Burchfiel AND Shuran Song}, 
    TITLE     = {{Diffusion Policy: Visuomotor Policy Learning via Action Diffusion}}, 
    BOOKTITLE = {Proceedings of Robotics: Science and Systems}, 
    YEAR      = {2023}, 
    ADDRESS   = {Daegu, Republic of Korea}, 
    MONTH     = {July}, 
    DOI       = {10.15607/RSS.2023.XIX.026} 
}

@article{kim24openvla,
    title={OpenVLA: An Open-Source Vision-Language-Action Model},
    author={{Moo Jin} Kim and Karl Pertsch and Siddharth Karamcheti and Ted Xiao and Ashwin Balakrishna and Suraj Nair and Rafael Rafailov and Ethan Foster and Grace Lam and Pannag Sanketi and Quan Vuong and Thomas Kollar and Benjamin Burchfiel and Russ Tedrake and Dorsa Sadigh and Sergey Levine and Percy Liang and Chelsea Finn},
    journal = {arXiv preprint arXiv:2406.09246},
    year={2024}
}

@article{shi2025diversity,
  title={Is Diversity All You Need for Scalable Robotic Manipulation?},
  author={Shi, Modi and Chen, Li and Chen, Jin and Lu, Yuxiang and Liu, Chiming and Ren, Guanghui and Luo, Ping and Huang, Di and Yao, Maoqing and Li, Hongyang},
  journal={arXiv preprint arXiv:2507.06219},
  year={2025}
}

@inproceedings{jiang2024dexmimicen,
      title     = {DexMimicGen: Automated Data Generation for Bimanual Dexterous Manipulation via Imitation Learning},
      author    = {Jiang, Zhenyu and Xie, Yuqi and Lin, Kevin and Xu, Zhenjia and Wan, Weikang and Mandlekar, Ajay and Fan, Linxi and Zhu, Yuke},
      booktitle = {2025 IEEE International Conference on Robotics and Automation (ICRA)},
      year      = {2025}
}

@inproceedings{chenobject,
      title={Object-Centric Dexterous Manipulation from Human Motion Data},
      author={Chen, Yuanpei and Wang, Chen and Yang, Yaodong and Liu, Karen},
      booktitle={8th Annual Conference on Robot Learning},
year      = {2024}
    }

@misc{lyu2024scissorbotlearninggeneralizablescissor,
      title={ScissorBot: Learning Generalizable Scissor Skill for Paper Cutting via Simulation, Imitation, and Sim2Real}, 
      author={Jiangran Lyu and Yuxing Chen and Tao Du and Feng Zhu and Huiquan Liu and Yizhou Wang and He Wang},
      year={2024},
      eprint={2409.13966},
      archivePrefix={arXiv},
      primaryClass={cs.RO},
      url={https://arxiv.org/abs/2409.13966}, 
}

@ARTICLE{9158349,
  author={Kadian, Abhishek and Truong, Joanne and Gokaslan, Aaron and Clegg, Alexander and Wijmans, Erik and Lee, Stefan and Savva, Manolis and Chernova, Sonia and Batra, Dhruv},
  journal={IEEE Robotics and Automation Letters}, 
  title={Sim2Real Predictivity: Does Evaluation in Simulation Predict Real-World Performance?}, 
  year={2020},
  volume={5},
  number={4},
  pages={6670-6677},
  doi={10.1109/LRA.2020.3013848}}

@INPROCEEDINGS{10610566,
  author={Katara, Pushkal and Xian, Zhou and Fragkiadaki, Katerina},
  booktitle={2024 IEEE International Conference on Robotics and Automation (ICRA)}, 
  title={Gen2Sim: Scaling up Robot Learning in Simulation with Generative Models}, 
  year={2024},
  volume={},
  number={},
  pages={6672-6679},
  doi={10.1109/ICRA57147.2024.10610566}}

@misc{jia2025discoverseefficientrobotsimulation,
      title={DISCOVERSE: Efficient Robot Simulation in Complex High-Fidelity Environments}, 
      author={Yufei Jia and Guangyu Wang and Yuhang Dong and Junzhe Wu and Yupei Zeng and Haonan Lin and Zifan Wang and Haizhou Ge and Weibin Gu and Kairui Ding and Zike Yan and Yunjie Cheng and Yue Li and Ziming Wang and Chuxuan Li and Wei Sui and Lu Shi and Guanzhong Tian and Ruqi Huang and Guyue Zhou},
      year={2025},
      eprint={2507.21981},
      archivePrefix={arXiv},
      primaryClass={cs.RO},
      url={https://arxiv.org/abs/2507.21981}, 
}

@article{deng2025graspvla,
    title={GraspVLA: a Grasping Foundation Model Pre-trained on Billion-scale Synthetic Action Data}, 
    author={Shengliang Deng and Mi Yan and Songlin Wei and Haixin Ma and Yuxin Yang and Jiayi Chen and Zhiqi Zhang and Taoyu Yang and Xuheng Zhang and Wenhao Zhang and Heming Cui and Zhizheng Zhang and He Wang},
    year={2025},
    eprint={2505.03233},
    archivePrefix={arXiv},
    primaryClass={cs.RO},
    url={https://arxiv.org/abs/2505.03233}
}

@misc{jain2025polarisscalablerealtosimevaluations,
      title={PolaRiS: Scalable Real-to-Sim Evaluations for Generalist Robot Policies}, 
      author={Arhan Jain and Mingtong Zhang and Kanav Arora and William Chen and Marcel Torne and Muhammad Zubair Irshad and Sergey Zakharov and Yue Wang and Sergey Levine and Chelsea Finn and Wei-Chiu Ma and Dhruv Shah and Abhishek Gupta and Karl Pertsch},
      year={2025},
      eprint={2512.16881},
      archivePrefix={arXiv},
      primaryClass={cs.RO},
      url={https://arxiv.org/abs/2512.16881}, 
}

@article{doi:10.1177/02783649251396980,
author = {Kejun Hu and Peng Yu and Ning Tan},
title ={Learning high-fidelity robot self-model with articulated 3D Gaussian splatting},
journal = {The International Journal of Robotics Research},
volume = {0},
number = {0},
year = {2025},
doi = {10.1177/02783649251396980},
URL = { 
    https://doi.org/10.1177/02783649251396980
},
eprint = { 
    https://doi.org/10.1177/02783649251396980
}
}

@misc{li2024evaluatingrealworldrobotmanipulation,
      title={Evaluating Real-World Robot Manipulation Policies in Simulation}, 
      author={Xuanlin Li and Kyle Hsu and Jiayuan Gu and Karl Pertsch and Oier Mees and Homer Rich Walke and Chuyuan Fu and Ishikaa Lunawat and Isabel Sieh and Sean Kirmani and Sergey Levine and Jiajun Wu and Chelsea Finn and Hao Su and Quan Vuong and Ted Xiao},
      year={2024},
      eprint={2405.05941},
      archivePrefix={arXiv},
      primaryClass={cs.RO},
      url={https://arxiv.org/abs/2405.05941}, 
}

@article{guo2025ctrl,
  title={Ctrl-world: A controllable generative world model for robot manipulation},
  author={Guo, Yanjiang and Shi, Lucy Xiaoyang and Chen, Jianyu and Finn, Chelsea},
  journal={arXiv preprint arXiv:2510.10125},
  year={2025}
}

@misc{geng2025roboverse,
      title={RoboVerse: Towards a Unified Platform, Dataset and Benchmark for Scalable and Generalizable Robot Learning}, 
      author={Haoran Geng and Feishi Wang and Songlin Wei and Yuyang Li and Bangjun Wang and Boshi An and Charlie Tianyue Cheng and Haozhe Lou and Peihao Li and Yen-Jen Wang and Yutong Liang and Dylan Goetting and Chaoyi Xu and Haozhe Chen and Yuxi Qian and Yiran Geng and Jiageng Mao and Weikang Wan and Mingtong Zhang and Jiangran Lyu and Siheng Zhao and Jiazhao Zhang and Jialiang Zhang and Chengyang Zhao and Haoran Lu and Yufei Ding and Ran Gong and Yuran Wang and Yuxuan Kuang and Ruihai Wu and Baoxiong Jia and Carlo Sferrazza and Hao Dong and Siyuan Huang and Yue Wang and Jitendra Malik and Pieter Abbeel},
      year={2025},
      eprint={2504.18904},
      archivePrefix={arXiv},
      primaryClass={cs.RO},
      url={https://arxiv.org/abs/2504.18904}, 
}

@misc{khazatsky2025droidlargescaleinthewildrobot,
      title={DROID: A Large-Scale In-The-Wild Robot Manipulation Dataset}, 
      author={Alexander Khazatsky and Karl Pertsch and Suraj Nair and Ashwin Balakrishna and Sudeep Dasari and Siddharth Karamcheti and Soroush Nasiriany and Mohan Kumar Srirama and Lawrence Yunliang Chen and Kirsty Ellis and Peter David Fagan and Joey Hejna and Masha Itkina and Marion Lepert and Yecheng Jason Ma and Patrick Tree Miller and Jimmy Wu and Suneel Belkhale and Shivin Dass and Huy Ha and Arhan Jain and Abraham Lee and Youngwoon Lee and Marius Memmel and Sungjae Park and Ilija Radosavovic and Kaiyuan Wang and Albert Zhan and Kevin Black and Cheng Chi and Kyle Beltran Hatch and Shan Lin and Jingpei Lu and Jean Mercat and Abdul Rehman and Pannag R Sanketi and Archit Sharma and Cody Simpson and Quan Vuong and Homer Rich Walke and Blake Wulfe and Ted Xiao and Jonathan Heewon Yang and Arefeh Yavary and Tony Z. Zhao and Christopher Agia and Rohan Baijal and Mateo Guaman Castro and Daphne Chen and Qiuyu Chen and Trinity Chung and Jaimyn Drake and Ethan Paul Foster and Jensen Gao and Vitor Guizilini and David Antonio Herrera and Minho Heo and Kyle Hsu and Jiaheng Hu and Muhammad Zubair Irshad and Donovon Jackson and Charlotte Le and Yunshuang Li and Kevin Lin and Roy Lin and Zehan Ma and Abhiram Maddukuri and Suvir Mirchandani and Daniel Morton and Tony Nguyen and Abigail O'Neill and Rosario Scalise and Derick Seale and Victor Son and Stephen Tian and Emi Tran and Andrew E. Wang and Yilin Wu and Annie Xie and Jingyun Yang and Patrick Yin and Yunchu Zhang and Osbert Bastani and Glen Berseth and Jeannette Bohg and Ken Goldberg and Abhinav Gupta and Abhishek Gupta and Dinesh Jayaraman and Joseph J Lim and Jitendra Malik and Roberto Martín-Martín and Subramanian Ramamoorthy and Dorsa Sadigh and Shuran Song and Jiajun Wu and Michael C. Yip and Yuke Zhu and Thomas Kollar and Sergey Levine and Chelsea Finn},
      year={2025},
      eprint={2403.12945},
      archivePrefix={arXiv},
      primaryClass={cs.RO},
      url={https://arxiv.org/abs/2403.12945}, 
}

@misc{dasari2020robonetlargescalemultirobotlearning,
      title={RoboNet: Large-Scale Multi-Robot Learning}, 
      author={Sudeep Dasari and Frederik Ebert and Stephen Tian and Suraj Nair and Bernadette Bucher and Karl Schmeckpeper and Siddharth Singh and Sergey Levine and Chelsea Finn},
      year={2020},
      eprint={1910.11215},
      archivePrefix={arXiv},
      primaryClass={cs.RO},
      url={https://arxiv.org/abs/1910.11215}, 
}

@inproceedings{
  fang2024rh20t,
  title        = {RH20T: A Comprehensive Robotic Dataset for Learning Diverse Skills in One-Shot},
  author       = {Fang, Hao-Shu and Fang, Hongjie and Tang, Zhenyu and Liu, Jirong and Wang, Chenxi and Wang, Junbo and Zhu, Haoyi and Lu, Cewu},
  booktitle    = {2024 IEEE International Conference on Robotics and Automation (ICRA)},
  pages        = {653--660},
  year         = {2024},
  organization = {IEEE}
}

@misc{embodimentcollaboration2025openxembodimentroboticlearning,
      title={Open X-Embodiment: Robotic Learning Datasets and RT-X Models}, 
      author={Embodiment Collaboration and Abby O'Neill and Abdul Rehman and Abhinav Gupta and Abhiram Maddukuri and Abhishek Gupta and Abhishek Padalkar and Abraham Lee and Acorn Pooley and Agrim Gupta and Ajay Mandlekar and Ajinkya Jain and Albert Tung and Alex Bewley and Alex Herzog and Alex Irpan and Alexander Khazatsky and Anant Rai and Anchit Gupta and Andrew Wang and Andrey Kolobov and Anikait Singh and Animesh Garg and Aniruddha Kembhavi and Annie Xie and Anthony Brohan and Antonin Raffin and Archit Sharma and Arefeh Yavary and Arhan Jain and Ashwin Balakrishna and Ayzaan Wahid and Ben Burgess-Limerick and Beomjoon Kim and Bernhard Schölkopf and Blake Wulfe and Brian Ichter and Cewu Lu and Charles Xu and Charlotte Le and Chelsea Finn and Chen Wang and Chenfeng Xu and Cheng Chi and Chenguang Huang and Christine Chan and Christopher Agia and Chuer Pan and Chuyuan Fu and Coline Devin and Danfei Xu and Daniel Morton and Danny Driess and Daphne Chen and Deepak Pathak and Dhruv Shah and Dieter Büchler and Dinesh Jayaraman and Dmitry Kalashnikov and Dorsa Sadigh and Edward Johns and Ethan Foster and Fangchen Liu and Federico Ceola and Fei Xia and Feiyu Zhao and Felipe Vieira Frujeri and Freek Stulp and Gaoyue Zhou and Gaurav S. Sukhatme and Gautam Salhotra and Ge Yan and Gilbert Feng and Giulio Schiavi and Glen Berseth and Gregory Kahn and Guangwen Yang and Guanzhi Wang and Hao Su and Hao-Shu Fang and Haochen Shi and Henghui Bao and Heni Ben Amor and Henrik I Christensen and Hiroki Furuta and Homanga Bharadhwaj and Homer Walke and Hongjie Fang and Huy Ha and Igor Mordatch and Ilija Radosavovic and Isabel Leal and Jacky Liang and Jad Abou-Chakra and Jaehyung Kim and Jaimyn Drake and Jan Peters and Jan Schneider and Jasmine Hsu and Jay Vakil and Jeannette Bohg and Jeffrey Bingham and Jeffrey Wu and Jensen Gao and Jiaheng Hu and Jiajun Wu and Jialin Wu and Jiankai Sun and Jianlan Luo and Jiayuan Gu and Jie Tan and Jihoon Oh and Jimmy Wu and Jingpei Lu and Jingyun Yang and Jitendra Malik and João Silvério and Joey Hejna and Jonathan Booher and Jonathan Tompson and Jonathan Yang and Jordi Salvador and Joseph J. Lim and Junhyek Han and Kaiyuan Wang and Kanishka Rao and Karl Pertsch and Karol Hausman and Keegan Go and Keerthana Gopalakrishnan and Ken Goldberg and Kendra Byrne and Kenneth Oslund and Kento Kawaharazuka and Kevin Black and Kevin Lin and Kevin Zhang and Kiana Ehsani and Kiran Lekkala and Kirsty Ellis and Krishan Rana and Krishnan Srinivasan and Kuan Fang and Kunal Pratap Singh and Kuo-Hao Zeng and Kyle Hatch and Kyle Hsu and Laurent Itti and Lawrence Yunliang Chen and Lerrel Pinto and Li Fei-Fei and Liam Tan and Linxi "Jim" Fan and Lionel Ott and Lisa Lee and Luca Weihs and Magnum Chen and Marion Lepert and Marius Memmel and Masayoshi Tomizuka and Masha Itkina and Mateo Guaman Castro and Max Spero and Maximilian Du and Michael Ahn and Michael C. Yip and Mingtong Zhang and Mingyu Ding and Minho Heo and Mohan Kumar Srirama and Mohit Sharma and Moo Jin Kim and Muhammad Zubair Irshad and Naoaki Kanazawa and Nicklas Hansen and Nicolas Heess and Nikhil J Joshi and Niko Suenderhauf and Ning Liu and Norman Di Palo and Nur Muhammad Mahi Shafiullah and Oier Mees and Oliver Kroemer and Osbert Bastani and Pannag R Sanketi and Patrick "Tree" Miller and Patrick Yin and Paul Wohlhart and Peng Xu and Peter David Fagan and Peter Mitrano and Pierre Sermanet and Pieter Abbeel and Priya Sundaresan and Qiuyu Chen and Quan Vuong and Rafael Rafailov and Ran Tian and Ria Doshi and Roberto Martín-Martín and Rohan Baijal and Rosario Scalise and Rose Hendrix and Roy Lin and Runjia Qian and Ruohan Zhang and Russell Mendonca and Rutav Shah and Ryan Hoque and Ryan Julian and Samuel Bustamante and Sean Kirmani and Sergey Levine and Shan Lin and Sherry Moore and Shikhar Bahl and Shivin Dass and Shubham Sonawani and Shubham Tulsiani and Shuran Song and Sichun Xu and Siddhant Haldar and Siddharth Karamcheti and Simeon Adebola and Simon Guist and Soroush Nasiriany and Stefan Schaal and Stefan Welker and Stephen Tian and Subramanian Ramamoorthy and Sudeep Dasari and Suneel Belkhale and Sungjae Park and Suraj Nair and Suvir Mirchandani and Takayuki Osa and Tanmay Gupta and Tatsuya Harada and Tatsuya Matsushima and Ted Xiao and Thomas Kollar and Tianhe Yu and Tianli Ding and Todor Davchev and Tony Z. Zhao and Travis Armstrong and Trevor Darrell and Trinity Chung and Vidhi Jain and Vikash Kumar and Vincent Vanhoucke and Vitor Guizilini and Wei Zhan and Wenxuan Zhou and Wolfram Burgard and Xi Chen and Xiangyu Chen and Xiaolong Wang and Xinghao Zhu and Xinyang Geng and Xiyuan Liu and Xu Liangwei and Xuanlin Li and Yansong Pang and Yao Lu and Yecheng Jason Ma and Yejin Kim and Yevgen Chebotar and Yifan Zhou and Yifeng Zhu and Yilin Wu and Ying Xu and Yixuan Wang and Yonatan Bisk and Yongqiang Dou and Yoonyoung Cho and Youngwoon Lee and Yuchen Cui and Yue Cao and Yueh-Hua Wu and Yujin Tang and Yuke Zhu and Yunchu Zhang and Yunfan Jiang and Yunshuang Li and Yunzhu Li and Yusuke Iwasawa and Yutaka Matsuo and Zehan Ma and Zhuo Xu and Zichen Jeff Cui and Zichen Zhang and Zipeng Fu and Zipeng Lin},
      year={2025},
      eprint={2310.08864},
      archivePrefix={arXiv},
      primaryClass={cs.RO},
      url={https://arxiv.org/abs/2310.08864}, 
}

@inproceedings{bu2025agibot_iros,
  title={Agibot world colosseo: A large-scale manipulation platform for scalable and intelligent embodied systems},
  author={Bu, Qingwen and Cai, Jisong and Chen, Li and Cui, Xiuqi and Ding, Yan and Feng, Siyuan and He, Xindong and Huang, Xu and others},
  booktitle={2025 IEEE/RSJ International Conference on Intelligent Robots and Systems (IROS)},
  year={2025},
  organization={IEEE}
}

@inproceedings{robocasa2024,
  title={RoboCasa: Large-Scale Simulation of Everyday Tasks for Generalist Robots},
  author={Soroush Nasiriany and Abhiram Maddukuri and Lance Zhang and Adeet Parikh and Aaron Lo and Abhishek Joshi and Ajay Mandlekar and Yuke Zhu},
  booktitle={Robotics: Science and Systems},
  year={2024}
}

@inproceedings{zhang2024dexgraspnet,
  title={DexGraspNet 2.0: Learning Generative Dexterous Grasping in Large-scale Synthetic Cluttered Scenes},
  author={Zhang, Jialiang and Liu, Haoran and Li, Danshi and Yu, XinQiang and Geng, Haoran and Ding, Yufei and Chen, Jiayi and Wang, He},
  booktitle={8th Annual Conference on Robot Learning},
  year={2024}
}

@article{chen2025robotwin,
  title={Robotwin 2.0: A scalable data generator and benchmark with strong domain randomization for robust bimanual robotic manipulation},
  author={Chen, Tianxing and Chen, Zanxin and Chen, Baijun and Cai, Zijian and Liu, Yibin and Li, Zixuan and Liang, Qiwei and Lin, Xianliang and Ge, Yiheng and Gu, Zhenyu and others},
  journal={arXiv preprint arXiv:2506.18088},
  year={2025}
}

@inproceedings{yu2019meta,
  title={Meta-World: A Benchmark and Evaluation for Multi-Task and Meta Reinforcement Learning},
  author={Tianhe Yu and Deirdre Quillen and Zhanpeng He and Ryan Julian and Karol Hausman and Chelsea Finn and Sergey Levine},
  booktitle={Conference on Robot Learning (CoRL)},
  year={2019},
  eprint={1910.10897},
  archivePrefix={arXiv},
  primaryClass={cs.LG},
  url={https://arxiv.org/abs/1910.10897}
}

@article{jing2025humanoidgen,
        title={HumanoidGen: Data Generation for Bimanual Dexterous Manipulation via LLM Reasoning},
        author={Jing, Zhi and Yang, Siyuan and Ao, Jicong and Xiao, Ting and Jiang, Yugang and Bai, Chenjia},
        journal={arXiv preprint arXiv:2507.00833},
        year={2025}
      }

@misc{sferrazza2024humanoidbench,
    title={HumanoidBench: Simulated Humanoid Benchmark for Whole-Body Locomotion and Manipulation},
    author={Carmelo Sferrazza and Dun-Ming Huang and Xingyu Lin and Youngwoon Lee and Pieter Abbeel},
    year={2024},
}

@article{chernyadev2024bigym,
  title={BiGym: A Demo-Driven Mobile Bi-Manual Manipulation Benchmark},
  author={Chernyadev, Nikita and Backshall, Nicholas and Ma, Xiao and Lu, Yunfan and Seo, Younggyo and James, Stephen},
  journal={arXiv preprint arXiv:2407.07788},
  year={2024}
}

@article{li2024behavior1k,
    title   = {BEHAVIOR-1K: A Human-Centered, Embodied AI Benchmark with 1,000 Everyday Activities and Realistic Simulation},
    author  = {Chengshu Li and Ruohan Zhang and Josiah Wong and Cem Gokmen and Sanjana Srivastava and Roberto Martín-Martín and Chen Wang and Gabrael Levine and Wensi Ai and Benjamin Martinez and Hang Yin and Michael Lingelbach and Minjune Hwang and Ayano Hiranaka and Sujay Garlanka and Arman Aydin and Sharon Lee and Jiankai Sun and Mona Anvari and Manasi Sharma and Dhruva Bansal and Samuel Hunter and Kyu-Young Kim and Alan Lou and Caleb R Matthews and Ivan Villa-Renteria and Jerry Huayang Tang and Claire Tang and Fei Xia and Yunzhu Li and Silvio Savarese and Hyowon Gweon and C. Karen Liu and Jiajun Wu and Li Fei-Fei},
    journal = {arXiv preprint arXiv:2403.09227},
    year    = {2024}
}

@inproceedings{ehsani2021manipulathor,
     title={ManipulaTHOR: A Framework for Visual Object Manipulation},
     author={Ehsani, Kiana and Han, Winson and Herrasti, Alvaro and VanderBilt, Eli and Weihs, Luca and Kolve, Eric and Kembhavi, Aniruddha and Mottaghi, Roozbeh},
     booktitle={CVPR},
     year={2021}
   }

@misc{yenamandra2024homerobotopenvocabularymobilemanipulation,
      title={HomeRobot: Open-Vocabulary Mobile Manipulation}, 
      author={Sriram Yenamandra and Arun Ramachandran and Karmesh Yadav and Austin Wang and Mukul Khanna and Theophile Gervet and Tsung-Yen Yang and Vidhi Jain and Alexander William Clegg and John Turner and Zsolt Kira and Manolis Savva and Angel Chang and Devendra Singh Chaplot and Dhruv Batra and Roozbeh Mottaghi and Yonatan Bisk and Chris Paxton},
      year={2024},
      eprint={2306.11565},
      archivePrefix={arXiv},
      primaryClass={cs.RO},
      url={https://arxiv.org/abs/2306.11565}, 
}

@inproceedings{chen2023daxbench,
    title={DaXBench: Benchmarking Deformable Object Manipulation with Differentiable Physics},
    author={Siwei Chen* and Yiqing Xu* and Cunjun Yu* and Linfeng Li and Xiao Ma and Zhongwen Xu and David Hsu},
    year={2023},
    booktitle={ICLR}
}

@inproceedings{corl2020softgym,
 title={SoftGym: Benchmarking Deep Reinforcement Learning for Deformable Object Manipulation},
 author={Lin, Xingyu and Wang, Yufei and Olkin, Jake and Held, David},
 booktitle={Conference on Robot Learning},
 year={2020}
}

@misc{huang2021plasticinelabsoftbodymanipulationbenchmark,
      title={PlasticineLab: A Soft-Body Manipulation Benchmark with Differentiable Physics}, 
      author={Zhiao Huang and Yuanming Hu and Tao Du and Siyuan Zhou and Hao Su and Joshua B. Tenenbaum and Chuang Gan},
      year={2021},
      eprint={2104.03311},
      archivePrefix={arXiv},
      primaryClass={cs.LG},
      url={https://arxiv.org/abs/2104.03311}, 
}

@misc{fang2025intentionexecutionprobinggeneralization,
      title={From Intention to Execution: Probing the Generalization Boundaries of Vision-Language-Action Models}, 
      author={Irving Fang and Juexiao Zhang and Shengbang Tong and Chen Feng},
      year={2025},
      eprint={2506.09930},
      archivePrefix={arXiv},
      primaryClass={cs.RO},
      url={https://arxiv.org/abs/2506.09930}, 
}

@misc{luu2025manifeelbenchmarkingunderstandingvisuotactile,
      title={ManiFeel: Benchmarking and Understanding Visuotactile Manipulation Policy Learning}, 
      author={Quan Khanh Luu and Pokuang Zhou and Zhengtong Xu and Zhiyuan Zhang and Qiang Qiu and Yu She},
      year={2025},
      eprint={2505.18472},
      archivePrefix={arXiv},
      primaryClass={cs.RO},
      url={https://arxiv.org/abs/2505.18472}, 
}

@inproceedings{
zhou2025exploring,
    title={Exploring the Limits of Vision-Language-Action Manipulation in Cross-task Generalization},
    author={Jiaming Zhou and Ke Ye and Jiayi Liu and Teli Ma and Zifan Wang and Ronghe Qiu and Kun-Yu Lin and Zhilin Zhao and Junwei Liang},
    booktitle={The Thirty-ninth Annual Conference on Neural Information Processing Systems},
    year={2025},
    url={https://openreview.net/forum?id=h6xQClTm4W}
}

@misc{zhang2025scenelanguagerepresentingscenes,
      title={The Scene Language: Representing Scenes with Programs, Words, and Embeddings}, 
      author={Yunzhi Zhang and Zizhang Li and Matt Zhou and Shangzhe Wu and Jiajun Wu},
      year={2025},
      eprint={2410.16770},
      archivePrefix={arXiv},
      primaryClass={cs.CV},
      url={https://arxiv.org/abs/2410.16770}, 
}

@article{DBLP:journals/corr/abs-1912-13470,
  author       = {Haoshu Fang and
                  Chenxi Wang and
                  Minghao Gou and
                  Cewu Lu},
  title        = {GraspNet: {A} Large-Scale Clustered and Densely Annotated Datase for
                  Object Grasping},
  journal      = {CoRR},
  volume       = {abs/1912.13470},
  year         = {2019},
  url          = {http://arxiv.org/abs/1912.13470},
  eprinttype    = {arXiv},
  eprint       = {1912.13470},
  bibsource    = {dblp computer science bibliography, https://dblp.org}
}

@article{sundaralingam2023curobo,
  title={curobo: Parallelized collision-free minimum-jerk robot motion generation},
  author={Sundaralingam, Balakumar and Hari, Siva Kumar Sastry and Fishman, Adam and Garrett, Caelan and Van Wyk, Karl and Blukis, Valts and Millane, Alexander and Oleynikova, Helen and Handa, Ankur and Ramos, Fabio and others},
  journal={arXiv preprint arXiv:2310.17274},
  year={2023}
}

@misc{intelligence2025pi05visionlanguageactionmodelopenworld,
      title={$\pi_{0.5}$: a Vision-Language-Action Model with Open-World Generalization}, 
      author={Physical Intelligence and Kevin Black and Noah Brown and James Darpinian and Karan Dhabalia and Danny Driess and Adnan Esmail and Michael Equi and Chelsea Finn and Niccolo Fusai and Manuel Y. Galliker and Dibya Ghosh and Lachy Groom and Karol Hausman and Brian Ichter and Szymon Jakubczak and Tim Jones and Liyiming Ke and Devin LeBlanc and Sergey Levine and Adrian Li-Bell and Mohith Mothukuri and Suraj Nair and Karl Pertsch and Allen Z. Ren and Lucy Xiaoyang Shi and Laura Smith and Jost Tobias Springenberg and Kyle Stachowicz and James Tanner and Quan Vuong and Homer Walke and Anna Walling and Haohuan Wang and Lili Yu and Ury Zhilinsky},
      year={2025},
      eprint={2504.16054},
      archivePrefix={arXiv},
      primaryClass={cs.LG},
      url={https://arxiv.org/abs/2504.16054}, 
}

@article{chen2024pgsr,
  title={PGSR: Planar-based Gaussian Splatting for Efficient and High-Fidelity Surface Reconstruction},
  author={Chen, Danpeng and Li, Hai and Ye, Weicai and Wang, Yifan and Xie, Weijian and Zhai, Shangjin and Wang, Nan and Liu, Haomin and Bao, Hujun and Zhang, Guofeng},
  journal={arXiv preprint arXiv:2406.06521},
  year={2024}
}

@inproceedings{wu2025difix3d+,
  title={DIFIX3D+: Improving 3D Reconstructions with Single-Step Diffusion Models},
  author={Wu, Jay Zhangjie and Zhang, Yuxuan and Turki, Haithem and Ren, Xuanchi and Gao, Jun and Shou, Mike Zheng and Fidler, Sanja and Gojcic, Zan and Ling, Huan},
  booktitle={Proceedings of the Computer Vision and Pattern Recognition Conference},
  pages={26024--26035},
  year={2025}
}

@article{bu2025univla,
  title={Univla: Learning to act anywhere with task-centric latent actions},
  author={Bu, Qingwen and Yang, Yanting and Cai, Jisong and Gao, Shenyuan and Ren, Guanghui and Yao, Maoqing and Luo, Ping and Li, Hongyang},
  journal={arXiv preprint arXiv:2505.06111},
  year={2025}
}

@article{liu2024rdt,
  title={RDT-1B: a Diffusion Foundation Model for Bimanual Manipulation},
  author={Liu, Songming and Wu, Lingxuan and Li, Bangguo and Tan, Hengkai and Chen, Huayu and Wang, Zhengyi and Xu, Ke and Su, Hang and Zhu, Jun},
  journal={arXiv preprint arXiv:2410.07864},
  year={2024}
}

@article{zheng2025x,
  title   = {X-VLA: Soft-Prompted Transformer as Scalable Cross-Embodiment Vision-Language-Action Model},
  author  = {Zheng, Jinliang and Li, Jianxiong and Wang, Zhihao and Liu, Dongxiu and Kang, Xirui
             and Feng, Yuchun and Zheng, Yinan and Zou, Jiayin and Chen, Yilun and Zeng, Jia and others},
  journal = {arXiv preprint arXiv:2510.10274},
  year    = {2025}
}

@inproceedings{bai2024colmap,
  title={Colmap-pcd: An open-source tool for fine image-to-point cloud registration},
  author={Bai, Chunge and Fu, Ruijie and Gao, Xiang},
  booktitle={2024 IEEE International Conference on Robotics and Automation (ICRA)},
  pages={1723--1729},
  year={2024},
  organization={IEEE}
}

@article{kerbl20233d,
  title={3D Gaussian splatting for real-time radiance field rendering.},
  author={Kerbl, Bernhard and Kopanas, Georgios and Leimk{\"u}hler, Thomas and Drettakis, George},
  journal={ACM Trans. Graph.},
  volume={42},
  number={4},
  pages={139--1},
  year={2023}
}

@article{ye2025gsplat,
  title={gsplat: An open-source library for Gaussian splatting},
  author={Ye, Vickie and Li, Ruilong and Kerr, Justin and Turkulainen, Matias and Yi, Brent and Pan, Zhuoyang and Seiskari, Otto and Ye, Jianbo and Hu, Jeffrey and Tancik, Matthew and others},
  journal={Journal of Machine Learning Research},
  volume={26},
  number={34},
  pages={1--17},
  year={2025}
}

@inproceedings{detone2018superpoint,
  title={Superpoint: Self-supervised interest point detection and description},
  author={DeTone, Daniel and Malisiewicz, Tomasz and Rabinovich, Andrew},
  booktitle={Proceedings of the IEEE conference on computer vision and pattern recognition workshops},
  pages={224--236},
  year={2018}
}

@inproceedings{lindenberger2023lightglue,
  title={Lightglue: Local feature matching at light speed},
  author={Lindenberger, Philipp and Sarlin, Paul-Edouard and Pollefeys, Marc},
  booktitle={Proceedings of the IEEE/CVF international conference on computer vision},
  pages={17627--17638},
  year={2023}
}

@inproceedings{dong2015domain,
  title={Domain-size pooling in local descriptors: DSP-SIFT},
  author={Dong, Jingming and Soatto, Stefano},
  booktitle={Proceedings of the IEEE conference on computer vision and pattern recognition},
  pages={5097--5106},
  year={2015}
}

@inproceedings{gr00tn1_2025,
  archivePrefix = {arxiv},
  eprint     = {2503.14734},
  title      = {{GR00T} {N1}: An Open Foundation Model for Generalist Humanoid Robots},
  author     = {NVIDIA and Johan Bjorck and Fernando Castañeda, Nikita Cherniadev and Xingye Da and Runyu Ding and Linxi "Jim" Fan and Yu Fang and Dieter Fox and Fengyuan Hu and Spencer Huang and Joel Jang and Zhenyu Jiang and Jan Kautz and Kaushil Kundalia and Lawrence Lao and Zhiqi Li and Zongyu Lin and Kevin Lin and Guilin Liu and Edith Llontop and Loic Magne and Ajay Mandlekar and Avnish Narayan and Soroush Nasiriany and Scott Reed and You Liang Tan and Guanzhi Wang and Zu Wang and Jing Wang and Qi Wang and Jiannan Xiang and Yuqi Xie and Yinzhen Xu and Zhenjia Xu and Seonghyeon Ye and Zhiding Yu and Ao Zhang and Hao Zhang and Yizhou Zhao and Ruijie Zheng and Yuke Zhu},
  month      = {March},
  year       = {2025},
  booktitle  = {ArXiv Preprint},
}

@misc{black2026pi0visionlanguageactionflowmodel,
      title={$\pi_0$: A Vision-Language-Action Flow Model for General Robot Control}, 
      author={Kevin Black and Noah Brown and Danny Driess and Adnan Esmail and Michael Equi and Chelsea Finn and Niccolo Fusai and Lachy Groom and Karol Hausman and Brian Ichter and Szymon Jakubczak and Tim Jones and Liyiming Ke and Sergey Levine and Adrian Li-Bell and Mohith Mothukuri and Suraj Nair and Karl Pertsch and Lucy Xiaoyang Shi and James Tanner and Quan Vuong and Anna Walling and Haohuan Wang and Ury Zhilinsky},
      year={2026},
      eprint={2410.24164},
      archivePrefix={arXiv},
      primaryClass={cs.LG},
      url={https://arxiv.org/abs/2410.24164}, 
}

@misc{zhong2026acotvlaactionchainofthoughtvisionlanguageaction,
      title={ACoT-VLA: Action Chain-of-Thought for Vision-Language-Action Models}, 
      author={Linqing Zhong and Yi Liu and Yifei Wei and Ziyu Xiong and Maoqing Yao and Si Liu and Guanghui Ren},
      year={2026},
      eprint={2601.11404},
      archivePrefix={arXiv},
      primaryClass={cs.RO},
      url={https://arxiv.org/abs/2601.11404}, 
}

\end{document}